\documentclass[10pt,a4paper]{article}
\usepackage[
  margin=18mm,
  top=25mm,
  headheight=41pt,
  headsep=8pt
]{geometry}
\usepackage[authoryear,round]{natbib}
\setcitestyle{semicolon,aysep={,},yysep={,}}
\usepackage{etoolbox}
\usepackage{booktabs,multirow,mathtools}
\usepackage{algorithm,algpseudocode}
\usepackage{subcaption}
\usepackage{placeins}
\usepackage{stfloats}
\usepackage{titlesec}
\usepackage{adjustbox}
\usepackage{microtype}
\titleformat{\section}{\large\bfseries}{\thesection}{0.6em}{}
\titleformat{\subsection}{\normalsize\bfseries}{\thesubsection}{0.6em}{}

\usepackage{styles/lab-preprint}
\newcommand{\PaperTitle}{Affordance-Conditioned Decision Making: Bridging the Semantic-Spatial Gap in Zero-Shot Cross-Floor Vision-and-Language Navigation}

\newcommand{\PaperAuthors}{%
  Xuekang Yang\textsuperscript{1},
  Lu Chen\textsuperscript{1},
  Shuang Luo\textsuperscript{1},
  Jialing Zhu\textsuperscript{2}\\[3pt]
  Qi Zhang\textsuperscript{3,*},
  Yue Gao\textsuperscript{4,*},
  Xiang Zhang\textsuperscript{3,*}%
}

\newcommand{\PaperAffiliation}{%
  \parbox{0.95\textwidth}{\centering
    \textsuperscript{1}School of Computer Science,
    Shanghai Jiao Tong University\\[3pt]
    \textsuperscript{2}School of Automation and Intelligent Sensing,
    Shanghai Jiao Tong University\\[3pt]
    \textsuperscript{3}Defense Innovation Institute,
    Academy of Military Sciences\\[3pt]
    \textsuperscript{4}MoE Key Laboratory of Artificial Intelligence
    and AI Institute, Shanghai Jiao Tong University\\[5pt]
    \textsuperscript{*}Corresponding authors
  }%
}

\newcommand{\PaperContact}{}

\renewcommand{\LabName}{RL\textsuperscript{2} Lab}
\renewcommand{\LabLogoPath}{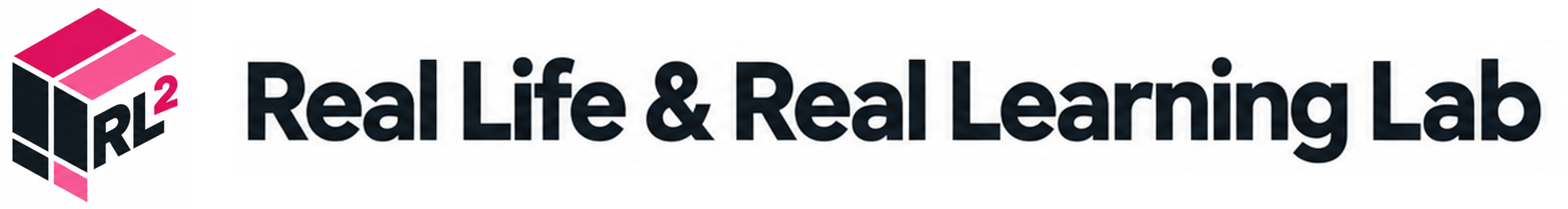}

\hypersetup{
  pdftitle={\PaperTitle},
  pdfauthor={Xuekang Yang, Lu Chen, Shuang Luo, Jialing Zhu,
    Qi Zhang, Yue Gao, Xiang Zhang}
}

\begin{document}

\title{\bfseries\PaperTitle}
\author{%
  \PaperAuthors\\[4pt]
  \small\PaperAffiliation\\
  \small\texttt{\PaperContact}
}
\date{}

\twocolumn[
  \begin{@twocolumnfalse}
    \maketitle
    \begin{abstract}
      Vision-and-language navigation increasingly relies on general-purpose semantic planners, yet translating correct high-level intent into reliable physical execution remains difficult in spatially constrained transitions. Reaching a staircase, doorway, or narrow passage does not ensure traversal; the agent must identify an executable affordance pose and recover from accumulated action errors. We propose PACE (\textbf{P}reference-refined \textbf{A}ffordance-\textbf{C}onditioned \textbf{E}xecution), a supervised local execution module that augments frozen zero-shot semantic planners for reliable cross-floor navigation. PACE grounds transition-related semantics into a long-horizon, agent-centric traversable affordance pose and conditions short-horizon action generation on this spatial target, thereby aligning semantic goals with physical execution. We further post-train PACE through failure-aware preference refinement using rollout-derived pairs that contrast normal or recovery behaviors with deviation-amplifying behaviors, thereby improving closed-loop correction. We integrate PACE into six open-source zero-shot VLN navigators and demonstrate consistent improvements on the cross-floor subsets of R2R-CE and RxR-CE, increasing the average success rate from 16.35\% to 27.65\% and from 4.76\% to 12.06\%, respectively. Real-world experiments further demonstrate PACE’s applicability in unseen environments, highlighting the potential of traversable affordances to bridge semantic intent and reliable embodied behavior.

    \end{abstract}
    \vspace{8pt}
  \end{@twocolumnfalse}
]
\thispagestyle{labfirst}

\section{Introduction}

Vision-and-language navigation in continuous environments requires
embodied agents to translate natural-language instructions into
low-level motion~\citep{10.1007/978-3-030-58604-1_7}. Recent zero-shot
and learning-based methods, increasingly driven by general-purpose
semantic planners, have made substantial
progress~\citep{zheng2026threestepnavhierarchicalgloballocal,
chen2025constraintawarezeroshotvisionlanguagenavigation,he2026strider,
wei2025ground,chu2026abot,
zhang2026qwenrobotworldtechnicalreportunifying}. Nevertheless, multi-floor environments expose a persistent challenge: correct semantic intent does not ensure reliable execution at constrained
transitions.

\begin{figure*}[!t]
    \centering
    \includegraphics[width=\textwidth]{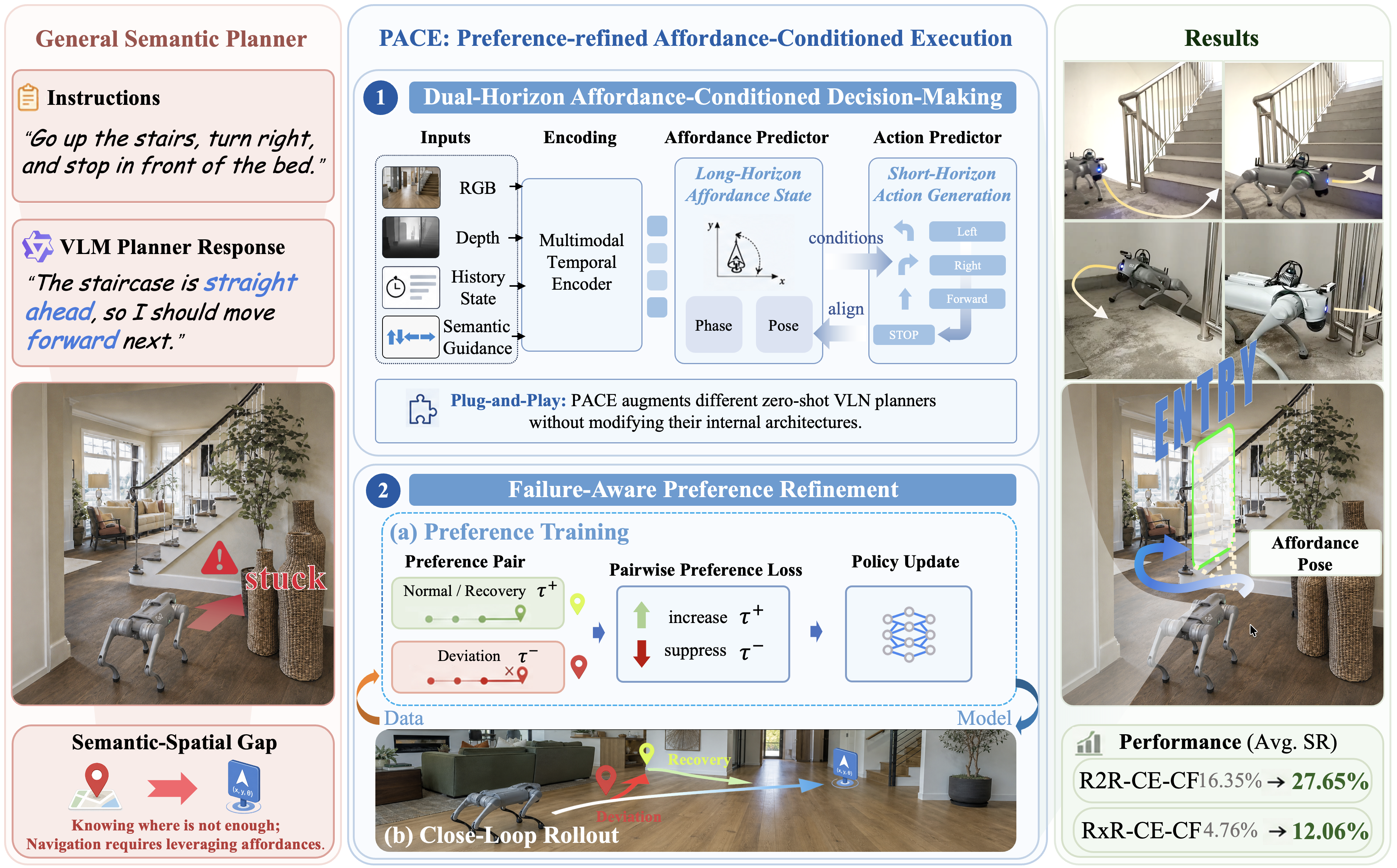}
    \caption{\textbf{Overview of PACE.}
PACE bridges the semantic-spatial gap through traversable affordance
estimation, affordance-conditioned action generation, and rollout-derived
preference refinement.}
    \label{fig:framework}
\end{figure*}

Existing cross-floor navigation methods primarily rely on explicit
geometric representations, including hierarchical scene
graphs~\citep{werby2024hierarchical}, floor-aware
maps~\citep{11128607,11358651,aerr}, and reconstructed traversable
surfaces~\citep{li2026bridging,zheng2026travexplorer}, to model
inter-floor connectivity and support geometric planning. However, stair
traversal remains a major source of failure: nearly 21\% of navigation
failures have been associated with unsuccessful stair
traversal~\citep{ramrakhya2023pirlnav}, while MANSION reports substantial
degradation on multi-floor long-horizon tasks~\citep{mansion}. These
findings indicate that geometry can describe multi-floor structure and
localize transition regions, but reaching a staircase does not guarantee
successful traversal.

Recent work has characterized a related challenge as the
\emph{semantic-geometric gap}, namely grounding semantic concepts in
environment-centric 3D structures~\citep{li2026bridging}. Despite strong
semantic perception, VLMs remain limited in reasoning over 3D structures
and action-induced spatial
transitions~\citep{yang2025thinking,pmlr-v267-chen25cr,
windecker2025navitrace,kong2025autospatial}. The semantic-geometric gap
specifies \emph{where} a relevant structure is, whereas reliable
navigation additionally requires an agent-centric executable affordance
pose specifying where and how the agent should approach and traverse it.
We term this remaining disconnect the \emph{semantic-spatial gap}. These results show that geometric localization of a staircase does not ensure its successful traversal.

Existing methods address this gap through two main paradigms.
Hierarchical methods insert waypoint-based spatial subgoals between
semantic reasoning and motion
execution~\citep{an2024etpnav,qiao2025open,shi2025smartway}. However,
these waypoints encode where the agent \emph{can} move rather than the
instruction-conditioned pose from which a transition \emph{should} be
executed, while external planners provide limited semantic conditioning
during execution~\citep{wang2026semantic,li2026bridging}. End-to-end
imitation-learning policies instead map visual-language observations
directly to actions~\citep{wei2025ground,chu2026abot,zhang2024navid,
10.1007/978-3-030-58604-1_7}, but expert-state-dominated supervision
leaves them vulnerable to compounding errors and limits recovery from
policy-induced deviations~\citep{ross2011reduction,shao2024offline}.
Thus, neither paradigm jointly provides task-conditioned executable
spatial grounding and robust closed-loop execution beyond the
expert-state distribution.

To address these limitations, we propose PACE
(\textbf{P}reference-refined \textbf{A}ffordance-\textbf{C}onditioned
\textbf{E}xecution) in Fig.~\ref{fig:framework}, a plug-and-play supervised
local execution framework trained on the corresponding benchmark training
split to augment frozen zero-shot semantic planners for cross-floor VLN;
both components remain frozen in unseen evaluation environments.
Our contributions are:

\begin{itemize}
    \item We introduce PACE, a dual-horizon
    affordance-conditioned framework that grounds stair-transition
    semantics into long-horizon traversable affordance poses for
    conditioning short-horizon action generation.

    \item We propose failure-aware preference refinement using
rollout-derived pairs, favoring normal and recovery behaviors over
deviation-amplifying alternatives to improve closed-loop correction
beyond expert demonstrations.

\item Across six open-source zero-shot VLN navigators, PACE improves
average SR by 11.30 and 7.30 percentage points on R2R-CE-CF and
RxR-CE-CF, respectively, and is successfully deployed in unseen real-world environments.
\end{itemize}

\section{Related Work}

\subsection{Semantic-Spatial Grounding}

Despite strong semantic understanding, multimodal large language models
(MLLMs) remain unreliable in spatial
reasoning~\citep{yang2025thinking,pmlr-v267-chen25cr,
windecker2025navitrace,kong2025autospatial}. Embodied navigation methods therefore combine vision-language and
segmentation models with depth and SLAM to associate semantics with 3D
structures represented by BEV or voxel maps, 3D Gaussians, and layered
memories~\citep{yu2026c,jang2026context,zhou2026beliefmapnav,
yokoyama2024vlfm,long2025instructnav,zhang2025mapnav,li2026bridging,
wang2026dynam3d,zheng20263dgsnavenhancingvisionlanguagemodel}.

However, these environment-centric representations primarily localize
semantic entities rather than model agent-centric executability.
Affordance-aware methods capture functional agent-environment interactions
but often depend on predefined skills or external intermediate
representations~\citep{pmlr-v205-ichter23a,yuan2024gamap,
chen2025affordances}. In contrast, PACE directly predicts agent-centric
traversable affordance poses from RGB-D observations and semantic guidance
without explicit 3D scene construction.

\subsection{Semantic-Aware Navigation}

Semantic-aware VLN methods mainly follow end-to-end and hierarchical
paradigms. End-to-end methods map visual-language observations to decisions
through task-specific policy
learning~\citep{wei2025ground,chu2026abot,zhang2024navid,
10.1007/978-3-030-58604-1_7} or zero-shot LLM/VLM
reasoning~\citep{zhou2024navgpt,bhatt2025vln,10611565}, whereas hierarchical
methods insert waypoints or topological nodes between semantic reasoning
and motion execution~\citep{an2024etpnav,qiao2025open,shi2025smartway}.

However, end-to-end policies learn semantic-to-action mappings implicitly,
zero-shot methods rely on unreliable long-horizon spatial reasoning, and
hierarchical subgoals typically encode geometric accessibility rather than
instruction-conditioned affordances~\citep{li2026bridging,
wang2026semantic}. PACE instead conditions local action generation on an
inferred affordance pose, explicitly grounding task semantics into
executable spatial decisions.

\section{Methods}
\paragraph{Problem Formulation}
Given an instruction $I$ and an unseen continuous environment
$\mathcal{E}$, the agent receives an egocentric RGB-D observation
$o_t=(r_t,d_t)$ at step $t$ and selects an action from
\begin{equation}
\mathcal{A}
=
\{\texttt{FORWARD},\texttt{LEFT},\texttt{RIGHT}\},
\end{equation}
where $r_t$ and $d_t$ denote the RGB and depth observations,
respectively. \texttt{FORWARD} advances the agent by $0.25\,\mathrm{m}$,
while \texttt{LEFT} and \texttt{RIGHT} rotate it by $30^\circ$. We focus
on cross-floor episodes requiring stair traversal.

The high-level VLM provides traversal guidance
$s_t\in\{\textsc{Up},\textsc{Down},\textsc{Unknown}\}$. PACE operates
on a $K$-step context
$c_t=(o_{t-K+1:t},\xi_{t-K+1:t},s_t)$, where
$\xi_i=(a_{i-1},\hat{p}^{\mathrm{aff}}_{i-1},\hat{\phi}_{i-1})$
contains the preceding action, affordance pose, and traversal phase.
PACE predicts
\begin{equation}
\begin{aligned}
h_t^{\mathrm{aff}}&=f_{\mathrm{aff}}(c_t),
\qquad \hat{z}_t=(\hat{p}^{\mathrm{aff}}_t,\hat{\phi}_t),\\
\hat{p}^{\mathrm{aff}}_t&=(\hat{x}_t,\hat{y}_t,\hat{\theta}_t),
\end{aligned}
\end{equation}
where $(\hat{x}_t,\hat{y}_t,\hat{\theta}_t)$ denotes the agent-centric
traversable affordance pose---the desired entry pose before traversal
and exit pose once traversal begins---and
$\hat{\phi}_t\in\allowbreak
\{\textsc{Approach},\allowbreak\textsc{Entry},\allowbreak\textsc{Traverse},\allowbreak\textsc{Exit}\}$
denotes the traversal phase.

PACE then generates a short-horizon action proposal
\begin{equation}
\hat{\tau}_t
=
f_{\mathrm{act}}(c_t,h_t^{\mathrm{aff}})
=
(\hat{a}_{t,1},\ldots,\hat{a}_{t,M_t},\texttt{STOP}),
\end{equation}
where $\hat{a}_{t,j}\in\mathcal{A}$ for $j\leq M_t$,
$\hat{a}_{t,M_t+1}\equiv\texttt{STOP}$, and $M_t$ counts the motion
actions preceding the terminal token; hence
$|\hat{\tau}_t|=M_t+1$. The \texttt{STOP} token terminates only the
current proposal and triggers the next closed-loop decision rather than
ending the navigation episode. Throughout this paper, zero-shot refers to the high-level navigator and
test-time protocol: PACE is supervised on the corresponding training split,
while both PACE and the navigator remain frozen without supervision or
adaptation in the unseen evaluation environments.
\subsection{Framework Overview}

As shown in Fig.~\ref{fig:framework}, PACE combines dual-horizon
affordance-conditioned decision making with failure-aware preference
refinement. Imitation learning couples long-horizon affordance estimation
with short-horizon action generation, while preference post-training on
rollout-derived deviation and recovery trajectories improves closed-loop
correction. At inference, the frozen module augments existing VLM
planners for local stair traversal.

\subsection{Dual-Horizon Affordance-Conditioned Execution}

\paragraph{Modeling.}
For compact notation, let
$z_t=(p_t^{\mathrm{aff}},\phi_t)$ denote the long-horizon affordance
state, where
$p_t^{\mathrm{aff}}=(x_t,y_t,\theta_t)$ is an agent-centric traversable
affordance pose and $\phi_t$ is the current traversal phase. PACE first
extracts an affordance-aware latent representation:
\begin{equation}
h_t^{\mathrm{aff}}
=
f_{\mathrm{aff}}(c_t;\omega),
\label{eq:affordance_representation}
\end{equation}
where $\omega$ denotes the parameters of the multimodal context encoder
and long-horizon affordance estimator. PACE parameterizes the
dual-horizon policy as
\begin{equation}
p_{\Theta}(z_t,\tau_t\mid c_t)
=
p_{\omega}(z_t\mid h_t^{\mathrm{aff}})
p_{\psi}(\tau_t\mid c_t,h_t^{\mathrm{aff}}),
\label{eq:dual_horizon}
\end{equation}
where $\psi$ denotes the parameters of the action generator and
$\tau_t=(a_{t,1},\ldots,a_{t,M_t},\texttt{STOP})$
denotes the short-horizon action sequence, with
$a_{t,j}\in\mathcal{A}$ for $j\leq M_t$ and
$a_{t,M_t+1}\equiv\texttt{STOP}$.

The long-horizon affordance state is decoded through parallel pose and
phase heads:
\begin{equation}
\begin{gathered}
\hat{p}_t^{\mathrm{aff}}
=
f_{\mathrm{pose}}(h_t^{\mathrm{aff}};\omega),
\\
p_{\omega}(\phi\mid h_t^{\mathrm{aff}})
=
f_{\mathrm{phase}}(h_t^{\mathrm{aff}};\omega),
\\
\hat{\phi}_t
=
\arg\max_{\phi}
p_{\omega}(\phi\mid h_t^{\mathrm{aff}}).
\end{gathered}
\label{eq:affordance_estimation}
\end{equation}
yielding
$\hat{z}_t=(\hat{p}_t^{\mathrm{aff}},\hat{\phi}_t)$.
Conditioned on $h_t^{\mathrm{aff}}$, the action generator
autoregressively models
\begin{equation}
p_{\psi}(\tau_t\mid c_t,h_t^{\mathrm{aff}})
=
\prod_{j=1}^{|\tau_t|}
p_{\psi}
\left(
a_{t,j}
\mid
a_{t,<j},
c_t,
h_t^{\mathrm{aff}}
\right).
\label{eq:action_factorization}
\end{equation}
Here $a_{t,<j}=(a_{t,1},\ldots,a_{t,j-1})$ denotes the preceding
action prefix.
Expert-path height trends delimit the four traversal phases:
$p_t^{\mathrm{aff},*}$ targets the entry pose during
\textsc{Approach} and the exit pose during
\textsc{Entry}/\textsc{Traverse}, while \textsc{Exit} is terminal.
The selected pose is projected into the current agent frame using its
horizontal displacement and tangent heading.

Given an expert dataset
$\mathcal{D}=\{(c_t,p_t^{\mathrm{aff}*},\phi_t^*,\tau_t^*)\}$
constructed from expert stair-traversal trajectories, where
$p_t^{\mathrm{aff}*}$ is the phase-dependent entry or exit pose,
$\phi_t^*$ is the traversal-phase label, and $\tau_t^*$ is the expert
short-horizon action sequence, we define
{\small
\begin{equation}
\mathcal{L}_{\mathrm{dual}}(\Theta;\mathcal{D})
=
\mathbb{E}_{\mathcal{D}}
\left[
\begin{aligned}
&\phantom{-}
\underbrace{
\lambda_p
\mathcal{L}_{\mathrm{pose}}
\left(
\hat{p}_t^{\mathrm{aff}},
p_t^{\mathrm{aff}*}
\right)
}_{\text{(1) Affordance pose regression}}
\\[-0.5mm]
&-
\underbrace{
\lambda_{\phi}
\log p_{\omega}
\left(
\phi_t^*
\mid
h_t^{\mathrm{aff}}
\right)
}_{\text{(2) Traversal phase classification}}
\\[-0.5mm]
&-
\underbrace{
\lambda_a
\sum_{j=1}^{|\tau_t^*|}
\log p_{\psi}
\left(
a_{t,j}^*
\mid
a_{t,<j}^*,
c_t,
h_t^{\mathrm{aff}}
\right)
}_{\text{(3) Action sequence classification}}
\end{aligned}
\right].
\label{eq:dual_horizon_objective}
\end{equation}
}

The supervised initialization is obtained by
\[
\Theta^{(0)}
=
\arg\min_{\Theta}
\mathcal{L}_{\mathrm{dual}}(\Theta;\mathcal{D}),
\]
where $\Theta=\{\omega,\psi\}$ denotes all trainable parameters, and
$\lambda_p$, $\lambda_{\phi}$, and $\lambda_a$ balance the three
supervision terms. Terms (1)--(3) correspond to traversable affordance
pose regression, traversal phase classification, and autoregressive
action prediction, respectively. Conditioning action generation on
$h_t^{\mathrm{aff}}$ couples long-horizon affordance grounding with
short-horizon physical execution.

\paragraph{Network Architecture.}
\begin{figure}[!htbp]
    \centering
    \includegraphics[width=\linewidth]{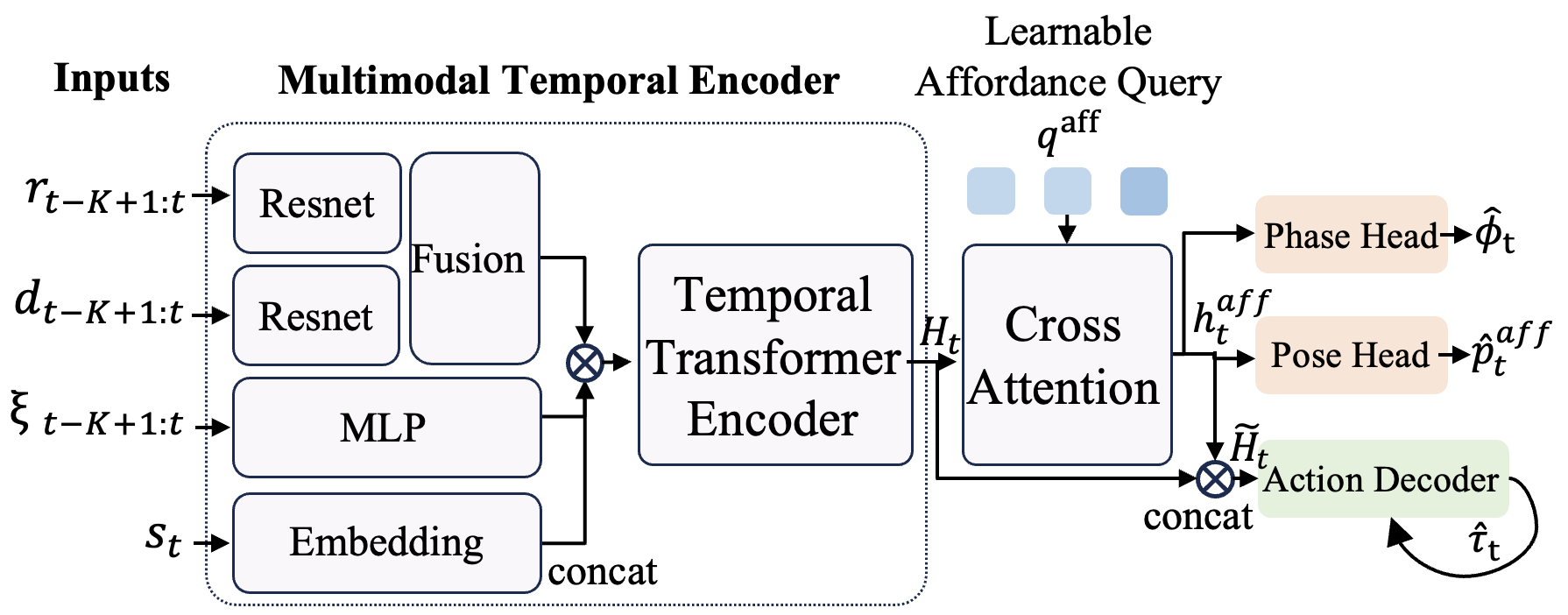}
    \caption{Model Architecture.}
    \label{fig:model_architecture}
\end{figure}
As illustrated in Fig.~\ref{fig:model_architecture}, PACE comprises a
multimodal temporal encoder, a query-based affordance estimator, and a
Transformer-based causal action decoder. RGB observations
$r_{t-K+1:t}$ and depth observations $d_{t-K+1:t}$ are encoded by
separate ImageNet-pretrained ResNet backbones and projected into a shared
$d$-dimensional feature space, where a gated fusion module integrates the
two visual streams. The historical context $\xi_{t-K+1:t}$ is encoded
by an MLP, while the semantic guidance $s_t$ is represented by a
learnable embedding. The resulting visual, historical, and semantic
features are concatenated into one multimodal token per context step.
After adding temporal positional embeddings, a Transformer encoder
models dependencies across the $K$-step context and produces
$H_t\in\mathbb{R}^{K\times d}$, where $d$ denotes the hidden dimension.

A learnable affordance query
$q^{\mathrm{aff}}\in\mathbb{R}^{d}$ performs multi-head
cross-attention over $H_t$ to extract
\begin{equation}
h_t^{\mathrm{aff}}
=
\operatorname{MHA}
\left(
q^{\mathrm{aff}},H_t,H_t
\right)
\in\mathbb{R}^{d}.
\end{equation}
Parallel phase and pose heads decode $h_t^{\mathrm{aff}}$ into the
traversal-phase prediction $\hat{\phi}_t$ and the traversable affordance
pose $\hat{p}_t^{\mathrm{aff}}$, respectively.

For short-horizon action generation, the affordance representation is
appended to the temporal features as an additional memory token:
\begin{equation}
\widetilde{H}_t
=
[H_t;h_t^{\mathrm{aff}}]
\in
\mathbb{R}^{(K+1)\times d}.
\label{eq:decoder_memory}
\end{equation}
A causal Transformer decoder parameterized by $\psi$ attends to
$\widetilde{H}_t$ and autoregressively predicts each action conditioned
on the preceding action prefix $a_{t,<j}$. The decoder is trained with
teacher forcing, while beam search generates candidate sequences at
inference until \texttt{STOP} or the maximum proposal length is reached.
The predicted affordance pose is further used for geometric reranking of
the candidate sequences.

\subsection{Failure-Aware Preference Refinement}

Imitation learning exposes PACE primarily to expert states, limiting
its ability to recover from policy-induced deviations during closed-loop
execution. We therefore refine the policy by favoring normal or recovery
behaviors over deviation-amplifying alternatives.

\paragraph{Preference Pair Construction.}
We collect model-induced states through autoregressive closed-loop
rollouts of the current PACE policy. A deviation is identified when the
positional or heading error from the expert route exceeds
$0.5\,\mathrm{m}$ or $30^\circ$ for two consecutive steps, whereas
failure is triggered immediately when either error exceeds
$1.5\,\mathrm{m}$ or $60^\circ$. From each failure point, A* plans a
trajectory back to the expert route, which is converted into discrete
expert actions and replayed under teacher forcing. Recovery ends when
the positional and heading errors remain within $0.25\,\mathrm{m}$ and
$30^\circ$ for two consecutive steps, after which subsequent states are
treated as normal behavior. We form fixed-length contiguous same-role snippets within each rollout and
match each deviation-amplifying snippet $\zeta^-$ with shuffled normal and
recovery snippets $\zeta^+$ from the global pools.

\paragraph{Preference Optimization.}
To distinguish a temporal preference snippet from the short-horizon
proposal $\tau_t$, we denote an ordered snippet as
\begin{equation}
\zeta
=
\bigl((c_i,\tau_i)\bigr)_{i=1}^{L},
\qquad
\tau_i
=
(a_{i,1},\ldots,a_{i,M_i},\texttt{STOP}),
\label{eq:preference_snippet}
\end{equation}
where $L$ is the snippet length, $M_i$ counts the motion actions preceding
\texttt{STOP}, and $|\tau_i|=M_i+1$ includes the terminal token. Given the affordance-aware
representation
$h_i^{\mathrm{aff}}=f_{\mathrm{aff}}(c_i;\omega)$,
each proposal is scored by its length-normalized conditional
log-likelihood:
\begin{equation}
\ell_{\Theta}(c_i,\tau_i)
=
\frac{1}{|\tau_i|}
\sum_{j=1}^{|\tau_i|}
\log p_{\psi}
\left(
a_{i,j}
\mid
a_{i,<j},
c_i,
h_i^{\mathrm{aff}}
\right),
\label{eq:proposal_preference_score}
\end{equation}
where $\Theta=\{\omega,\psi\}$. All proposals are rescored under
teacher forcing, with $a_{i,<j}$ denoting the recorded action prefix.
Recovery snippets contain A*-derived expert actions, whereas deviation
snippets contain model-generated actions collected during
autoregressive rollouts. Importantly, the scores are computed using
model-inferred rather than ground-truth affordance representations,
matching closed-loop inference. Length normalization prevents a bias
toward shorter proposals.

We aggregate the proposal scores over the temporal snippet:
\begin{equation}
S_{\Theta}(\zeta)
=
\frac{1}{L}
\sum_{i=1}^{L}
\ell_{\Theta}(c_i,\tau_i).
\label{eq:temporal_preference_score}
\end{equation}
Temporal aggregation captures whether consecutive decisions amplify
or reduce navigation errors, which cannot be reliably characterized
by isolated actions. Given a cross-rollout preference pair
$(\zeta^+,\zeta^-)$, we define
\begin{equation}
\begin{gathered}
\Delta_{\Theta}
=
S_{\Theta}(\zeta^+)-S_{\Theta}(\zeta^-),
\\
P_{\Theta}(\zeta^+\succ\zeta^-)
=
\sigma\left(\beta\Delta_{\Theta}\right).
\end{gathered}
\label{eq:preference_probability}
\end{equation}
where $\zeta^+$ denotes a normal or recovery snippet, $\zeta^-$ denotes
a deviation snippet, and $\beta$ controls the preference sharpness. The
pairwise preference loss is
\begin{equation}
\mathcal{L}_{\mathrm{pref}}
\left(
\Theta;\mathcal{D}_{\mathrm{pref}}
\right)
=
-
\mathbb{E}_{(\zeta^+,\zeta^-)\sim\mathcal{D}_{\mathrm{pref}}}
\left[
\log\sigma\left(\beta\Delta_{\Theta}\right)
\right].
\label{eq:preference_loss}
\end{equation}

The post-training objective is
\[
\begin{aligned}
\mathcal{L}_{\mathrm{post}}
&=\tfrac{1}{2}\mathcal{L}_{\mathrm{dual}}(\Theta;\mathcal{D}_{\mathrm{exp}})\\
&\quad+\tfrac{1}{2}\mathcal{L}_{\mathrm{dual}}(\Theta;\mathcal{D}_{\mathrm{on}})\\
&\quad+\lambda_{\mathrm{pref}}
\mathcal{L}_{\mathrm{pref}}(\Theta;\mathcal{D}_{\mathrm{pref}}).
\end{aligned}
\]
Equal-sized expert and on-policy batches preserve the original
affordance-estimation and action-generation capabilities while adapting
PACE to model-induced states, whereas the preference term favors normal
and recovery behaviors over deviation-amplifying alternatives.

\subsection{Training, Inference, and Plug-and-Play Integration}

Algorithm~\ref{alg:pace_training_inference} summarizes PACE training
and receding-horizon inference. Supervised initialization is followed
by iterative refinement using labeled on-policy states and
rollout-derived preference pairs.

\begin{algorithm}[t]
\caption{PACE Training and Inference}
\label{alg:pace_training_inference}
\small
\begin{algorithmic}[1]
\Require Expert data $\mathcal{D}_{\mathrm{exp}}$, refinement rounds $R$,
beam width $B$

\State $\Theta^{(0)}
\arg\min_{\Theta}
\mathcal{L}_{\mathrm{dual}}
(\Theta;\mathcal{D}_{\mathrm{exp}})$
\Comment{Imitation learning}

\For{$r=1,\ldots,R$}
    \State $\mathcal{T}^{(r)}
    \gets
    \operatorname{Rollout}
    (\pi_{\Theta^{(r-1)}})$
    \State $\mathcal{D}_{\mathrm{on}}^{(r)}
    \gets
    \operatorname{LabelOnPolicy}
    (\mathcal{T}^{(r)})$
    \State $(\zeta^{+},\zeta^{-})^{(r)}
    \gets
    \operatorname{BuildPreferencePairs}
    (\mathcal{T}^{(r)})$
    \State $\mathcal{D}_{\mathrm{pref}}^{(r)}
    \gets
    \{(\zeta^{+},\zeta^{-})^{(r)}\}$
    \State $\Theta^{(r)}
    \gets
    \arg\min_{\Theta}
    \mathcal{L}_{\mathrm{post}}
    (\Theta;
    \mathcal{D}_{\mathrm{exp}},
    \mathcal{D}_{\mathrm{on}}^{(r)},
    \mathcal{D}_{\mathrm{pref}}^{(r)})$
\EndFor

\Statex \textbf{Inference at step $t$:}
\State $\hat{z}_t
\gets
(\hat{p}^{\mathrm{aff}}_t,\hat{\phi}_t)$
\Comment{Affordance estimation}
\State $\mathcal{B}_t
\gets
\operatorname{BeamSearch}_{B}
(p_{\psi}(\tau_t|c_t,h_t^{\mathrm{aff}}))$
\State $\hat{\tau}_t
\gets
\arg\max_{\tau\in\mathcal{B}_t}
J_t(\tau)$
\Comment{Geometry-aware selection}
\State Execute $\hat{a}_{t,1}$ and replan

\end{algorithmic}
\end{algorithm}

At inference, beam search retains up to $B$ candidate short-horizon
action sequences:
\begin{equation}
\mathcal{B}_t
=
\operatorname{BeamSearch}_{B}
\left[
p_{\psi}
\left(
\tau_t
\mid
c_t,h_t^{\mathrm{aff}}
\right)
\right],
\qquad
|\mathcal{B}_t|\leq B.
\label{eq:beam_candidates}
\end{equation}
The candidates are rolled out with known motion primitives and reranked
by their likelihood and geometric consistency with the predicted
affordance pose:
\begin{equation}
\begin{aligned}
J_t(\tau)
&=
\log p_{\psi}
\left(
\tau\mid c_t,h_t^{\mathrm{aff}}
\right)
-
\lambda_g
G\left(
\tau,\hat{p}_t^{\mathrm{aff}}
\right),
\\
\hat{\tau}_t
&=
\arg\max_{\tau\in\mathcal{B}_t}
J_t(\tau),
\end{aligned}
\label{eq:geometry_reranking}
\end{equation}
Here $(r_{xy},r_\theta)
=\operatorname{Transform}(\hat{p}_t^{\mathrm{aff}};\tau)$ is the terminal
residual after executing $\tau$,
$G=\|r_{xy}\|_2+\lambda_\theta|\operatorname{wrap}(r_\theta)|
+\lambda_s\mathbf{1}[\texttt{STOP}\notin\tau]$, and
$\log p_\psi(\tau\mid c_t,h_t^{\mathrm{aff}})$ is the cumulative
sequence log-likelihood.

PACE interfaces with heterogeneous VLN agents through unified RGB-D,
semantic-guidance, and action interfaces, without accessing internal
representations or requiring retraining. GroundingDINO detects visible
staircases, while the VLM assesses their relevance and provides
directional guidance. PACE retains control until the phase head predicts \textsc{Exit} or a
safety/budget limit is reached, then returns control to the navigator and
may be reactivated if traversal remains incomplete.

\section{Experiments}
We evaluate PACE in terms of: (1) long-horizon affordance and short-horizon action prediction; (2) improvements to zero-shot VLN baselines; (3) real-world cross-floor navigation; and (4) standalone stair traversal and ablations of affordance conditioning and preference refinement.

\newcommand{\bootci}[2]{
  \multicolumn{2}{c}{\scriptsize$[#1,#2]$}
}

\newcommand{\sigup}{\ensuremath{^{\blacktriangle}}}
\newcommand{\unstable}{\ensuremath{^{\vartriangle}}}
\newcommand{\sigdrop}{\ensuremath{^{\blacktriangledown}}}

\begin{table*}[t]
\centering

{\small
\setlength{\tabcolsep}{1pt}
\renewcommand{\arraystretch}{1.05}

\begin{adjustbox}{max width=\linewidth}
\begin{tabular}{
    @{}l
    *{8}{r@{\,/\,}l}
    @{}
}
\toprule

&
\multicolumn{8}{c}{R2R-CE-CF}
& \multicolumn{8}{c}{RxR-CE-CF} \\
\cmidrule(lr){2-9}
\cmidrule(lr){10-17}

\multicolumn{1}{c}{Method}
& \multicolumn{2}{c}{SR (\%) $\uparrow$}
& \multicolumn{2}{c}{OSR (\%) $\uparrow$}
& \multicolumn{2}{c}{SPL (\%) $\uparrow$}
& \multicolumn{2}{c}{NDTW $\uparrow$}
& \multicolumn{2}{c}{SR (\%) $\uparrow$}
& \multicolumn{2}{c}{OSR (\%) $\uparrow$}
& \multicolumn{2}{c}{SPL (\%) $\uparrow$}
& \multicolumn{2}{c}{NDTW $\uparrow$} \\

\cmidrule(lr){2-3}
\cmidrule(lr){4-5}
\cmidrule(lr){6-7}
\cmidrule(lr){8-9}
\cmidrule(lr){10-11}
\cmidrule(lr){12-13}
\cmidrule(lr){14-15}
\cmidrule(lr){16-17}

&
Alone & +PACE
& Alone & +PACE
& Alone & +PACE
& Alone & +PACE
& Alone & +PACE
& Alone & +PACE
& Alone & +PACE
& Alone & +PACE \\
\midrule

VLN-Zero
& 6.90  & \textbf{23.75}\sigup
& 10.34 & \textbf{30.27}\sigup
& 6.00  & \textbf{18.45}\sigup
& 0.24  & \textbf{0.46}\sigup
& 3.23  & \textbf{7.37}\sigup
& 4.61  & \textbf{8.76}\sigup
& 2.14  & \textbf{4.87}\sigup
& 0.18  & \textbf{0.22}\sigup \\

3-step-Nav
& 6.90  & \textbf{19.54}\sigup
& 11.49 & \textbf{36.02}\sigup
& 5.95  & \textbf{12.53}\sigup
& 0.38  & \textbf{0.45}\sigup
& 5.99  & \textbf{10.14}\sigup
& 5.99  & \textbf{20.28}\sigup
& \textbf{4.81} & 3.65\unstable
& \textbf{0.37} & 0.32\sigdrop \\

CA-Nav
& 14.94 & \textbf{25.67}\sigup
& 27.97 & \textbf{56.70}\sigup
& 7.15  & \textbf{11.84}\sigup
& 0.17  & \textbf{0.23}\sigup
& 4.15  & \textbf{11.98}\sigup
& 11.98 & \textbf{25.35}\sigup
& 1.54  & \textbf{4.95}\sigup
& 0.17  & \textbf{0.20}\sigup \\

STRIDER
& 17.24 & \textbf{21.84}\unstable
& 21.84 & \textbf{29.50}\sigup
& 15.85 & \textbf{17.53}\unstable
& 0.47  & \textbf{0.51}\sigup
& 2.30  & \textbf{8.29}\sigup
& 4.15  & \textbf{11.98}\sigup
& 1.80  & \textbf{6.64}\sigup
& 0.35  & \textbf{0.38}\sigup \\

Open-Nav
& 20.31 & \textbf{24.90}\unstable
& 24.52 & \textbf{37.55}\sigup
& 18.43 & \textbf{19.99}\unstable
& 0.52  & \textbf{0.57}\sigup
& 3.23  & \textbf{6.91}\unstable
& 3.69  & \textbf{9.68}\sigup
& 2.84  & \textbf{5.76}\unstable
& 0.34  & \textbf{0.40}\sigup \\

HSGM
& 31.80 & \textbf{50.19}\sigup
& 42.91 & \textbf{63.22}\sigup
& 20.87 & \textbf{35.93}\sigup
& 0.29  & \textbf{0.34}\sigup
& 9.68  & \textbf{27.65}\sigup
& 13.82 & \textbf{37.79}\sigup
& 6.66  & \textbf{19.19}\sigup
& 0.38  & \textbf{0.46}\sigup \\

\bottomrule
\end{tabular}
\end{adjustbox}
}

\caption{
Cross-floor navigation results on the val-unseen splits of R2R-CE-CF
and RxR-CE-CF. Each pair reports Alone/+PACE, with the higher value in
bold. For the 95\% paired-bootstrap CI of
$\Delta=+\mathrm{PACE}-\mathrm{Alone}$,
$\blacktriangle$, $\vartriangle$, and $\blacktriangledown$ denote robust
improvement (CI${}>0$), inconclusive difference (CI contains zero), and
robust decrease (CI${}<0$), respectively.
}
\label{tab:main_cross_floor_results}
\end{table*}

\subsection{Experimental Setup}
\paragraph{Datasets.}
We evaluate on the cross-floor subsets of R2R-CE~\citep{10.1007/978-3-030-58604-1_7} and
RxR-CE(en-US)~\citep{11145171,ku2020room}, denoted R2R-CE-CF and RxR-CE-CF,
which retain episodes with start-to-goal vertical displacement
$|h_s-h_g|>1\,\mathrm{m}$. R2R-CE-CF contains 925 training and 261
val-unseen episodes, while the \texttt{en-US} subset of RxR-CE-CF
contains 932 and 217, respectively. PACE is trained on the corresponding
training splits, and all baselines are evaluated on the full val-unseen
splits.

\paragraph{Baselines and Metrics.}
We integrate PACE with six recent open-source zero-shot VLN systems.
VLN-Zero uses cache-enabled neurosymbolic planning~\citep{bhatt2025vln},
whereas CA-Nav performs constraint-aware value
mapping~\citep{chen2025constraintawarezeroshotvisionlanguagenavigation}.
3-step-Nav adopts global-local planning with trajectory
auditing~\citep{zheng2026threestepnavhierarchicalgloballocal}, while
Open-Nav employs open-source LLM reasoning~\citep{qiao2025open}.
STRIDER optimizes structured waypoint decisions~\citep{he2026strider},
whereas HSGM plans over hierarchical semantic-geometric
maps~\citep{li2026bridging}. Since prior works do not separately report cross-floor results, we
rerun all released baselines under a unified protocol using the same
high-level VLM, both alone and with PACE.

For overall cross-floor VLN, we report Success Rate (SR, within
$3\,\mathrm{m}$ of the goal), Oracle Success Rate (OSR), Success weighted
by Path Length (SPL), and Normalized Dynamic Time Warping (NDTW). For
local stair traversal, we report OSR, SPL, NDTW, and Collision Rate (CR).

\paragraph{Implementation Details.}
We locally deploy the open-source Qwen-3.6-27B-FP8~\citep{qwen2026qwen36}
as the shared high-level VLM for cost-efficient evaluation. PACE uses
five-step contexts and ImageNet-pretrained dual-stream ResNet-18
backbones. Supervised training runs for 15 epochs with AdamW, batch size
32, an initial learning rate of $10^{-4}$, and cosine decay.
Post-training performs 20 on-policy rounds at $10^{-5}$, collecting
2,048 rollouts per round and mixing equal-sized expert and on-policy
batches. Rollout starts combine recorded expert states and
GroundingDINO randomized approach states at a ratio of
$1{:}8$. Preference learning uses five-step snippets,
$\lambda_{\mathrm{pref}}=0.1$, and $\beta=1.0$. Inference uses beam size
5 and a maximum proposal length of 48. Training takes approximately
19 hours on four NVIDIA RTX 5090 GPUs. Real-world experiments use a Unitree Go2 Edu equipped with a RealSense D435i and a Livox Mid-360, with PACE inference running on an RTX 3070 Ti laptop.

\newcommand{\srrobust}[2]{
  \makebox[3.0em][r]{#1}\,/\,%
  \makebox[3.4em][l]{\textbf{#2}\ensuremath{^{\blacktriangle}}}
}

\newcommand{\srheader}{
  \makebox[3.0em][r]{Alone}\,/\,%
  \makebox[3.4em][l]{+PACE}
}

\begin{table}[!htbp]
\centering

{\fontsize{9}{10.8}\selectfont
\setlength{\tabcolsep}{1pt}
\renewcommand{\arraystretch}{1.08}

\begin{adjustbox}{max width=\linewidth}
\begin{tabular}{@{}l
  cc
  cc
  @{}
}
\toprule

\multirow[c]{2}{*}{\raisebox{-1.6ex}{Method}}
& \multicolumn{2}{c}{R2R-CE-CF}
& \multicolumn{2}{c}{RxR-CE-CF} \\

\cmidrule(lr){2-3}
\cmidrule(lr){4-5}

& \shortstack[c]{Takeover\\Rate (\%)}
& \shortstack[c]{SR (\%) $\uparrow$\\\srheader}
& \shortstack[c]{Takeover\\Rate (\%)}
& \shortstack[c]{SR (\%) $\uparrow$\\\srheader} \\

\midrule

VLN-Zero
& 39.08
& \srrobust{7.84}{51.96}
& 11.06
& \srrobust{8.33}{41.67} \\

3-step-Nav
& 51.72
& \srrobust{10.37}{37.04}
& 42.86
& \srrobust{10.75}{20.43} \\

CA-Nav
& 65.52
& \srrobust{19.30}{36.26}
& 35.48
& \srrobust{6.49}{28.57} \\

STRIDER
& 27.59
& \srrobust{19.44}{47.22}
& 17.05
& \srrobust{5.41}{24.32} \\

Open-Nav
& 38.31
& \srrobust{27.00}{41.00}
& 28.11
& \srrobust{8.20}{24.59} \\

HSGM
& 76.63
& \srrobust{34.50}{63.50}
& 51.15
& \srrobust{11.71}{52.25} \\

\bottomrule
\end{tabular}
\end{adjustbox}
}

\caption{
Takeover-subset SR, comparing Alone/+PACE on aligned episode IDs.
Takeover Rate is the fraction of episodes with actual PACE takeover;
$\blacktriangle$ denotes a paired-bootstrap 95\% CI above zero.
}
\label{tab:takeover_success_correlation}
\end{table}

\subsection{Main Results}

\begin{figure}[!htbp]
    \centering

    \begin{minipage}{\linewidth}
    \begin{subfigure}[t]{0.495\linewidth}
        \centering
        \includegraphics[width=\linewidth]{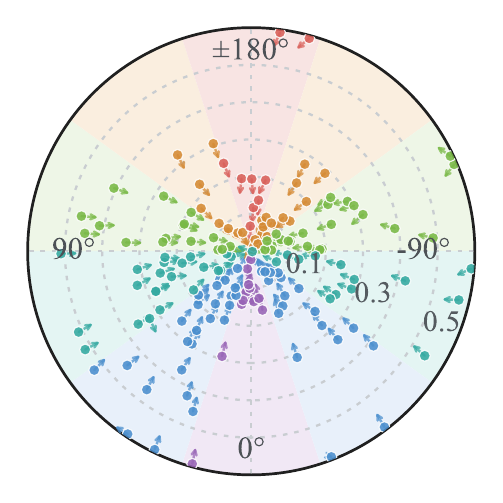}
        \caption{Affordance pose error.}
        \label{fig:pace_landmark}
    \end{subfigure}
    \hfill
    \begin{subfigure}[t]{0.485\linewidth}
        \centering
        \includegraphics[width=\linewidth]{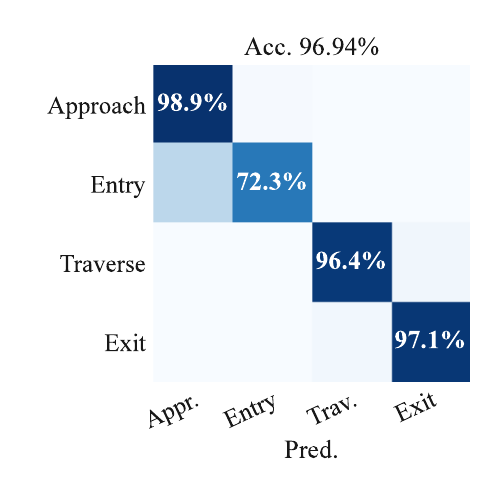}
        \caption{Phase prediction.}
        \label{fig:pace_phase}
    \end{subfigure}

    \vspace{0.4em}

    \begin{subfigure}[t]{0.485\linewidth}
        \centering
        \includegraphics[width=\linewidth]{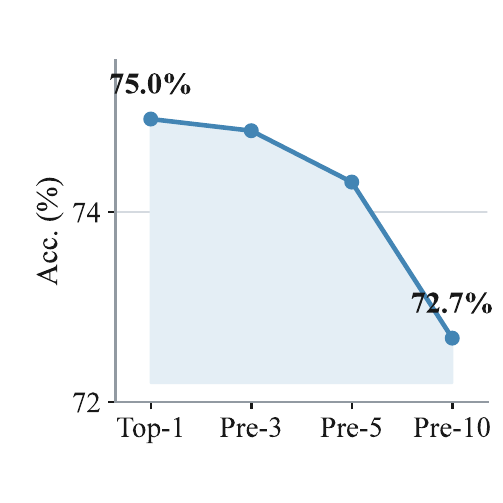}
        \caption{Action prefix accuracy.}
        \label{fig:pace_prefix}
    \end{subfigure}
    \hfill
    \begin{subfigure}[t]{0.485\linewidth}
        \centering
        \includegraphics[width=\linewidth]{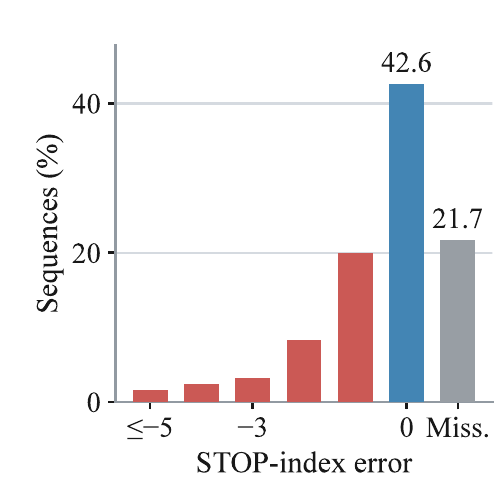}
        \caption{Termination error.}
        \label{fig:pace_termination}
    \end{subfigure}

    \end{minipage}

    \caption{
        Standalone PACE predictions on R2R-CE-CF val-unseen. In (a), the center denotes zero positional error; radius and angle encode its magnitude and direction, while arrow deviation from the inward direction represents heading error.
    }
    \label{fig:pace_prediction_diagnostics}
\end{figure}

\paragraph{PACE Prediction Performance.}
Figure~\ref{fig:pace_prediction_diagnostics} evaluates the standalone
predictive performance of PACE over both horizons through policy
rollouts on the R2R-CE-CF val-unseen split. For the long-horizon
affordance state, pose errors are concentrated near the origin, with a
mean positional error of $0.329\,\mathrm{m}$ and a yaw MAE of
$9.9^\circ$, indicating that PACE provides a reasonably accurate
spatial target for local execution. Phase prediction achieves an
accuracy of $96.94\%$, although $27.7\%$ of \textsc{Entry} states are
misclassified as \textsc{Approach}, revealing ambiguity at the boundary
between approaching and entering the stairs. For short-horizon action
generation, prefix accuracy decreases only modestly from $74.98\%$ for
the first action to $72.67\%$ over the first ten actions, suggesting
limited autoregressive error accumulation under rollout evaluation. Moreover, $42.6\%$ of \texttt{STOP} predictions are exact, and $71.0\%$ occur no more than two steps early. Overall, these rollout results demonstrate reliable affordance grounding
and stable action generation under policy-induced states, consistent
with the robustness objective of failure-aware preference refinement.

\paragraph{Quantitative Evaluation on VLN Benchmarks.}
Table~\ref{tab:main_cross_floor_results} shows that standalone zero-shot
VLN methods remain limited in cross-floor scenarios, with the best SR
reaching only 31.80\% on R2R-CE-CF and 9.68\% on RxR-CE-CF. With
PACE, all six methods improve across every metric on R2R-CE-CF, with
average gains of 11.30, 19.03, and 7.01 percentage points in SR, OSR,
and SPL, respectively, and 0.082 in NDTW. On RxR-CE-CF, PACE improves
SR and OSR for all methods and improves SPL and NDTW for five of six.
HSGM+PACE achieves the highest SR, OSR, and SPL on both benchmarks.

On episodes with actual PACE takeover,
Table~\ref{tab:takeover_success_correlation} shows SR gains of
9.68--44.12 percentage points across all navigator--benchmark pairs,
with all paired-bootstrap 95\% confidence intervals above zero.
However, non-random takeover and stochastic VLM sampling make this
analysis associative rather than causal. Across 10,000 paired-bootstrap
resamples of the full aligned results, PACE yields 40/48 robust
improvements, seven inconclusive differences, and one robust decrease.
Gain variations likely reflect takeover frequency and execution quality.
NDTW improves less because PACE targets critical stair transitions,
whereas NDTW evaluates the full trajectory. Overall, PACE is broadly
compatible with diverse zero-shot VLN navigators.

\paragraph{Qualitative Real-World Case Study.}
\begin{figure}[!htbp]
    \centering
    \includegraphics[width=0.92\linewidth]{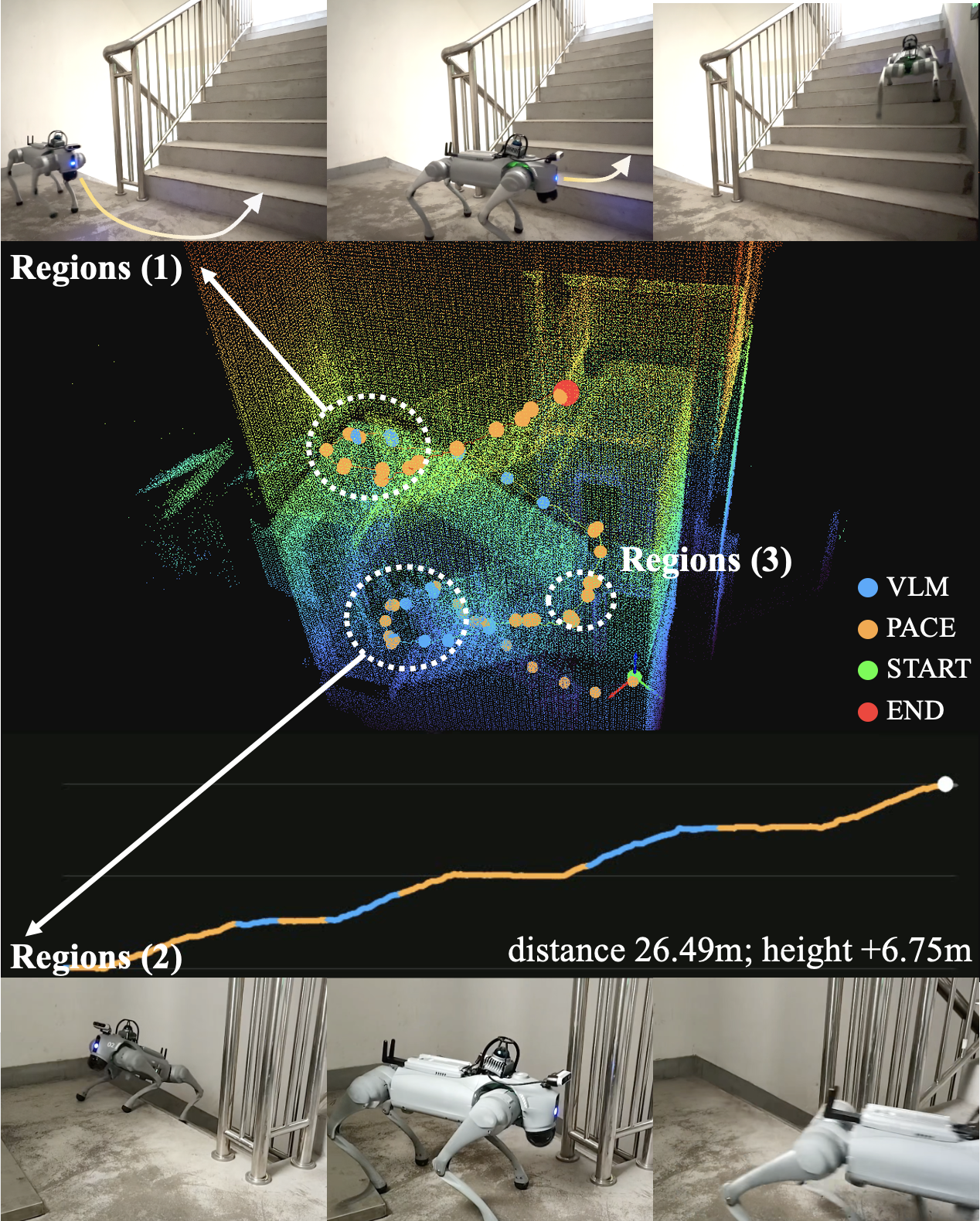}
    \caption{
Real-world two-floor stair traversal with VLN-Zero and PACE.
Blue and orange points denote VLM- and PACE-controlled trajectory
segments, respectively.
}
    \label{fig:real_world_stairpace}
\end{figure}

We deploy VLN-Zero~\citep{bhatt2025vln} with PACE without an explicit
navigation map or environment cache. Planning takes 0.2--0.5\,s for the
VLM or approximately 0.2\,s for PACE; with 2\,s MPC execution, each
control cycle takes 2.2--2.5\,s. As shown in
Fig.~\ref{fig:real_world_stairpace}, we evaluate the system in an unseen
two-floor stairwell with an elevation gain of approximately 7\,m under
the instruction \textit{``Enter the stairwell and proceed upstairs until
reaching the top floor.''} SLAM is used only for trajectory
visualization.

PACE successfully negotiates narrow and directionally ambiguous
stair-transition regions, as illustrated by Regions (1) and (2). At
stair corners, it predicts an executable affordance pose aligned with
the next flight and translates it into local actions, allowing the robot
to complete the turn and continue ascending. In Region (2), PACE
occasionally emits \texttt{STOP} before reaching the stair exit; the
high-level VLM detects the incomplete instruction and reactivates PACE,
enabling recovery and completion of the two-floor ascent. The alternating
control is reflected by the interleaved VLM- and PACE-controlled
trajectory segments.

PACE nevertheless exhibits local control instability during stair
traversal, including left--right oscillations and reduced clearance from
the railing. In Region (3), the clustered orange trajectory points
indicate repeated local corrections and reduced execution efficiency,
likely caused by sim-to-real viewpoint discrepancies and contact-induced
perturbations during ascent.

\subsection{Ablation Studies}

\begin{table}[!htbp]
\centering

{\small
\setlength{\tabcolsep}{1pt}
\renewcommand{\arraystretch}{1.05}

\begin{adjustbox}{max width=\linewidth}
\begin{tabular}{clcccc
  @{}
}
\toprule

\# & Method
& OSR (\%) $\uparrow$
& CR (\%) $\downarrow$
& SPL (\%) $\uparrow$
& NDTW (\%) $\uparrow$ \\
\midrule

1 & \textbf{PACE}
& \textbf{67.4}\,$\mathbf{\pm}$\,\textbf{1.1}
& \textbf{2.0}\,$\mathbf{\pm}$\,\textbf{0.2}
& \textbf{64.3}\,$\mathbf{\pm}$\,\textbf{2.9}
& \textbf{84.8}\,$\mathbf{\pm}$\,\textbf{1.1} \\

2 & \shortstack[l]{Waypoint Cond.}
& 63.1\,$\pm$\,0.5
& 11.5\,$\pm$\,4.5
& 59.9\,$\pm$\,2.1
& 70.6\,$\pm$\,1.0 \\

3 & \shortstack[l]{w/o Pref. Loss}
& 56.5\,$\pm$\,1.4
& 4.0\,$\pm$\,0.8
& 54.0\,$\pm$\,4.7
& 64.3\,$\pm$\,1.9 \\

4 & \shortstack[l]{w/o Aff. Cond.}
& 53.9\,$\pm$\,2.6
& 4.7\,$\pm$\,0.4
& 50.6\,$\pm$\,3.6
& 59.8\,$\pm$\,1.4 \\

5 & \shortstack[l]{Aux. Aff. Only}
& 49.3\,$\pm$\,3.1
& 2.9\,$\pm$\,0.2
& 47.6\,$\pm$\,2.8
& 57.3\,$\pm$\,3.5 \\

6 & \shortstack[l]{w/o F.A. Ref.}
& 37.3\,$\pm$\,0.9
& 8.3\,$\pm$\,2.2
& 34.5\,$\pm$\,4.0
& 57.4\,$\pm$\,1.3 \\

7 & \shortstack[l]{Action-Only BC}
& 17.5\,$\pm$\,0.5
& 9.2\,$\pm$\,3.7
& 15.6\,$\pm$\,6.9
& 51.7\,$\pm$\,2.6 \\

\bottomrule
\end{tabular}
\end{adjustbox}
}

\caption{
Standalone stair-traversal ablation of PACE. ``Aff.'', ``Cond.'',
``Pref.'', and ``F.A. Ref.'' denote affordance, conditioning,
preference, and failure-aware refinement, respectively. Results are
reported as mean${\pm}$standard deviation over three runs.
}
\label{tab:preference_refinement}
\end{table}

We conduct ablations on standalone stair-traversal segments of
R2R-CE-CF to isolate local execution from high-level VLM uncertainty
(Table~\ref{tab:preference_refinement}). Compared with waypoint
conditioning, PACE improves OSR by 4.3 percentage points while reducing
CR by 9.5 points and improving NDTW by 14.2 points, indicating that its
affordance state provides more effective guidance than a generic goal
pose. Removing affordance conditioning or using affordance prediction
only as auxiliary supervision decreases OSR by 13.5 and 18.1 points,
respectively, supporting its direct use by the action decoder. The
action-only BC baseline exhibits a 49.9-point OSR gap, confirming the
importance of the intermediate spatial representation. Removing the
preference loss reduces OSR by 10.9 points, whereas removing the entire
failure-aware refinement stage causes a 30.1-point drop, demonstrating
the complementary contributions of preference optimization and
policy-induced recovery supervision.

\section{Conclusion}

This work presents PACE, a plug-and-play affordance-conditioned
supervised local execution framework that connects zero-shot semantic
planners with reliable execution for cross-floor VLN. PACE grounds stair-transition
semantics into traversable affordance poses, conditions short-horizon
action generation on the learned affordance representation, and improves
closed-loop correction through failure-aware preference refinement.
Experiments across six zero-shot VLN navigators and real-world deployment
demonstrate its effectiveness and applicability. Future work will
extend PACE to other spatially constrained transitions, such as doorways
and narrow passages.

\FloatBarrier
\bibliographystyle{plainnat}
\bibliography{references}

\clearpage
\appendix
This document supplements the main submission. All notation, metrics,
baselines, and external models referred to below follow the main paper,
where the corresponding references are given.

\section{Failure-Aware Preference Refinement Details}

\paragraph{Rollout and recovery labels.}
Each refinement round collects 2,048 closed-loop rollouts from recorded
expert states and GroundingDINO-randomized approach states at a
$1{:}8$ ratio. The current policy is executed autoregressively, while
the privileged expert route is used only during training to compute
nearest-route positional and heading errors and to derive recovery
targets. A deviation begins when either error exceeds $0.5\,\mathrm{m}$
or $30^\circ$ for two consecutive steps. Failure is triggered
immediately when either error exceeds $1.5\,\mathrm{m}$ or $60^\circ$.
From a failure state, A* plans toward a future expert-route anchor, and
the resulting path is converted into the same $0.25\,\mathrm{m}$ and
$30^\circ$ motion primitives. Recovery ends when both errors remain
within $0.25\,\mathrm{m}$ and $30^\circ$ for two consecutive steps.
Algorithm~\ref{alg:recovery_labels} summarizes this training-time
labeling process.

\begin{algorithm}[!htbp]
\caption{Rollout Failure and Recovery Labeling}
\label{alg:recovery_labels}
\footnotesize
\begin{algorithmic}[1]
\Require Current policy $\pi_\Theta$ and privileged expert route
\State $m\gets\textsc{Normal}$; $n_{\rm drift}\gets0$;
$n_{\rm return}\gets0$
\While{the rollout is active}
  \State Compute nearest-route errors $(d_t,\delta_t)$
  \If{$m\neq\textsc{Recovery}$}
    \If{$d_t>0.5\,\mathrm{m}$ or $\delta_t>30^\circ$}
      \State $n_{\rm drift}\gets n_{\rm drift}+1$
    \Else
      \State $n_{\rm drift}\gets0$
    \EndIf
    \If{$n_{\rm drift}\geq2$}
      \State $m\gets\textsc{Deviation}$
    \EndIf
    \If{$d_t>1.5\,\mathrm{m}$ or $\delta_t>60^\circ$}
      \State Plan an A* path to a future expert-route anchor
      \State Discretize the path into recovery primitives
      \State $m\gets\textsc{Recovery}$
    \EndIf
  \Else
    \State Execute the next A*-derived recovery primitive
    \If{$d_t\leq0.25\,\mathrm{m}$ and $\delta_t\leq30^\circ$}
      \State $n_{\rm return}\gets n_{\rm return}+1$
    \Else
      \State $n_{\rm return}\gets0$
    \EndIf
    \If{$n_{\rm return}\geq2$}
      \State $m\gets\textsc{Normal}$
    \EndIf
  \EndIf
  \State Record the current state and proposal with role $m$
\EndWhile
\end{algorithmic}
\end{algorithm}
The expert route and A*-derived actions are privileged training signals
and are never available during evaluation.

\paragraph{Snippet pools and matching.}
Within each rollout, maximal contiguous runs with the same role
(\textsc{Normal}, \textsc{Deviation}, or \textsc{Recovery}) are divided
into all sliding windows of length five. The resulting windows are
pooled globally across rollouts within the current refinement round.
These pools are rebuilt, rather than accumulated, across rounds. After
the three pools are independently shuffled, each deviation window is
paired once with a normal window and once with a recovery window when
the corresponding positive pool is non-empty. Positive pools wrap
around when exhausted, producing at most two pairs per deviation
window without imposing an additional class ratio.

\begin{algorithm}[!htbp]
\caption{Preference-Pair Construction in Round $r$}
\label{alg:preference_pairs}
\scriptsize
\begin{algorithmic}[1]
\Require Policy $\pi_{\Theta^{r-1}}$ and expert data
\State $\mathcal{T}^{r}\gets
\operatorname{Rollout}(\pi_{\Theta^{r-1}})$
\State Label rollout states as normal, deviation, or recovery
\State Build length-5 pools $\mathcal{N}^{r}$,
$\mathcal{D}^{r}$, and $\mathcal{R}^{r}$
\State Independently shuffle $\mathcal{N}^{r}$,
$\mathcal{D}^{r}$, and $\mathcal{R}^{r}$
\State $\mathcal{P}^{r}\gets\varnothing$
\For{$k=1,\ldots,|\mathcal{D}^{r}|$}
  \If{$|\mathcal{N}^{r}|>0$}
    \State $u\gets1+((k-1)\bmod|\mathcal{N}^{r}|)$
    \State $\mathcal{P}^{r}\gets\mathcal{P}^{r}\cup
    \{(\mathcal{N}^{r}_{u},\mathcal{D}^{r}_{k})\}$
  \EndIf
  \If{$|\mathcal{R}^{r}|>0$}
    \State $v\gets1+((k-1)\bmod|\mathcal{R}^{r}|)$
    \State $\mathcal{P}^{r}\gets\mathcal{P}^{r}\cup
    \{(\mathcal{R}^{r}_{v},\mathcal{D}^{r}_{k})\}$
  \EndIf
\EndFor
\State Update $\Theta^{r}$ using equal expert/on-policy batches
and $\mathcal{P}^{r}$
\end{algorithmic}
\end{algorithm}
Every recorded proposal, including its terminal \texttt{STOP}, is
rescored under teacher forcing using model-predicted affordance
features. For a snippet
$\zeta=((c_i,\tau_i))_{i=1}^{L}$ with $L=5$, its score is
\begin{equation}
S_\Theta(\zeta)
=
\frac{1}{L}
\sum_{i=1}^{L}
\frac{1}{|\tau_i|}
\sum_{j=1}^{|\tau_i|}
\log p_\psi
\left(
a_{i,j}\mid
a_{i,<j},c_i,h_i^{\mathrm{aff}}
\right).
\label{eq:supp_preference_score}
\end{equation}
Token averaging removes a systematic preference for shorter proposals,
while temporal averaging evaluates whether consecutive decisions
sustain correction rather than relying on an isolated action.

\section{Implementation and Annotation Details}

\paragraph{Architecture and optimization.}
PACE uses separate ImageNet-pretrained ResNet-18 backbones for RGB and
depth. Their 512-dimensional outputs are each projected to
$d/2=128$ dimensions, which is the shared visual feature space referred
to in the main paper, and are combined by a learned gate. Historical
actions and phases use 16-dimensional embeddings, historical affordance
poses use a 64-dimensional MLP encoding, and directional guidance uses a
16-dimensional embedding. The concatenated visual, historical, and
semantic features form one context token per step, which is projected to
the $d=256$ Transformer hidden space. The temporal encoder has two
four-head layers, the query-based affordance estimator has one
four-head decoder layer, and the causal action decoder has two
four-head layers. All Transformer feed-forward layers have dimension
1,024. The complete model has 27.4M trainable parameters, including
22.4M in the two visual backbones.

\begin{table}[!htbp]
\centering
{\small
\setlength{\tabcolsep}{1pt}
\renewcommand{\arraystretch}{1.02}
\begin{adjustbox}{max width=\linewidth}
\begin{tabular}{ll@{}}
\toprule
Setting & Value \\
\midrule
Visual backbones & dual ImageNet-pretrained ResNet-18 \\
Context & five steps ($K=5$) \\
Hidden dimension & $d=256$ \\
Attention / FFN & 4 heads / 1,024 \\
Temporal/query/action & $2/1/2$  layers \\
Dropout & feature/Transformer $=0.35/0.20$ \\
Objective weights & $\lambda_p/\lambda_\phi/\lambda_a=1/0.5/1$ \\
Loss details & Smooth-L1 $\beta=1$; label smoothing 0.05 \\
Imitation learning & 15 epochs; batch 32 \\
IL optimizer & AdamW; LR $10^{-4}$; WD $5{\times}10^{-4}$ \\
Failure-aware refinement & 20 on-policy rounds; LR $10^{-5}$ \\
Refinement rollouts & 2,048 per round \\
Rollout starts & expert/randomized $=1{:}8$ \\
Refinement batches & expert/on-policy $=16/16$ \\
Preference & five-step snippets; $\lambda_{\rm pref}=0.1$; $\beta=1$ \\
Proposal & maximum length 48, including \texttt{STOP} \\
Inference & beam size 5 \\
Geom. weights & $(\lambda_g,\lambda_\theta,\lambda_s)=(0.35,0.25,0.5)$ \\
IL checkpoint & lowest validation objective \\
Refinement checkpoint & final refinement round \\
Seeds & 42 (main); 42/43/44 (ablations) \\
\bottomrule
\end{tabular}
\end{adjustbox}}
\caption{
PACE implementation settings. Imitation learning uses cosine
learning-rate decay; all test-time models are frozen.
}
\label{tab:supp_hyperparameters}
\end{table}

\paragraph{Cross-floor splits and scene isolation.}
We preserve the official R2R-CE and RxR-CE train/val-unseen assignments
and retain episodes with $|h_s-h_g|>1\,\mathrm{m}$. This yields 925/261
R2R-CE-CF and 932/217 RxR-CE-CF train/val-unseen episodes, matching the
main paper. The retained train/val-unseen episodes span 61/11 R2R scenes
and 59/11 RxR scenes, with no scene overlap between the two splits.

\paragraph{Pose and phase annotation.}
Algorithm~\ref{alg:label_generation} gives the deterministic label
construction. A non-flat height change starts a stair run, and the
preceding expert pose is treated as its entry. A direction reversal
closes the run immediately; alternatively, $F$ consecutive flat states
confirm a plateau, with the first flat state selected as the exit.
We use $F=3$ for R2R and $F=9$ for RxR and discard runs whose total
height change is below $0.5\,\mathrm{m}$. For runs with an observed
pre-stair pose, \textsc{Approach} denotes the retained prefix,
\textsc{Entry} marks the pre-stair key pose, \textsc{Traverse} covers
the interval toward the exit pose, and \textsc{Exit} is terminal. When
a sequence starts on the stairs, it enters \textsc{Traverse} directly
without an \textsc{Entry} label. Pose yaw follows the local expert-path
tangent.

Because \textsc{Exit} is terminal and defines no remaining traversal
target, no affordance pose label is assigned to it. The pose regression
term of $\mathcal{L}_{\mathrm{dual}}$ is therefore masked on
\textsc{Exit} states, and the expectation in that term is taken over the
non-terminal states only; the phase and action terms remain active on all
states. Equivalently, $\lambda_p$ is set to zero whenever
$\phi_t^*=\textsc{Exit}$.

\begin{algorithm}[!htbp]
\caption{Affordance Pose and Phase Labels}
\label{alg:label_generation}
\footnotesize
\begin{algorithmic}[1]
\Require Expert poses $\{q_i=(X_i,Y_i,Z_i,\theta_i)\}_{i=1}^{T}$
\State Quantize $\Delta Y_i$ as up/down/flat using $0.05\,\mathrm{m}$
\State Start a run at the first non-flat change; set entry $e=i-1$
\State Close at a sign reversal, or after $F$ flats using the first flat as exit $x$
\State Let $b\gets(e=1)$ indicate that the sequence starts on stairs
\If{$|Y_x-Y_e|\geq0.5\,\mathrm{m}$}
  \For{each index $j$ associated with this run}
    \If{$b$ and $j<x$}
      \State $\phi_j^*=\textsc{Traverse}$; $p_j^*=q_x$
    \ElsIf{$j<e$}
      \State $\phi_j^*=\textsc{Approach}$; $p_j^*=q_e$
    \ElsIf{$j=e$}
      \State $\phi_j^*=\textsc{Entry}$; $p_j^*=q_x$
    \ElsIf{$e<j<x$}
      \State $\phi_j^*=\textsc{Traverse}$; $p_j^*=q_x$
    \Else
      \State $\phi_j^*=\textsc{Exit}$; no pose target, mask
      $\mathcal{L}_{\mathrm{pose}}$
    \EndIf
  \EndFor
\EndIf
\end{algorithmic}
\end{algorithm}

For the current horizontal pose $(X_i,Z_i,\theta_i)$ and target
$(X^*,Z^*,\theta^*)$, let $\Delta X=X^*-X_i$ and
$\Delta Z=Z^*-Z_i$. The agent-centric label is
\begin{equation}
\begin{split}
x^* &=
\sin\theta_i\,\Delta X+\cos\theta_i\,\Delta Z,\\
y^* &=
-\cos\theta_i\,\Delta X+\sin\theta_i\,\Delta Z,\\
\theta_{\mathrm{rel}}^* &=
\operatorname{wrap}(\theta^*-\theta_i).
\end{split}
\label{eq:supp_agent_transform}
\end{equation}
Here, $+x$ points forward and $+y$ points left. Randomized approach
trajectories terminate at the same entry pose and use the same
agent-centric transformation.

\section{Integration Details}

\paragraph{Unified baseline adapter.}
PACE interfaces with all six frozen navigators through unified RGB-D,
semantic-guidance, and action interfaces. Each navigator retains its own
maps, memory, hidden states, and planner; none are exposed to PACE or
retrained. GroundingDINO first detects visible staircases. For each
detection, the VLM assesses its relevance to the current instruction and
returns a directional prediction
$d\in\{\textsc{Up},\textsc{Down},\textsc{Unknown}\}$ together with a
relevance confidence score $q$. PACE is activated when the VLM either
predicts an explicit traversal direction or confidently confirms that
the detected staircase is relevant to the instruction, i.e.,
$d\in\{\textsc{Up},\textsc{Down}\}$ or $q\geq\delta_{\mathrm{act}}$.
This confidence-based condition allows PACE to take over when the VLM
reliably recognizes the required staircase even if its directional
prediction remains \textsc{Unknown}. Otherwise, the navigator retains
control. Algorithm~\ref{alg:pace_takeover} summarizes this closed-loop
integration.

\begin{algorithm}[!htbp]
\caption{PACE Takeover and Handoff}
\label{alg:pace_takeover}
\footnotesize
\begin{algorithmic}[1]
\Require Frozen navigator $\mathcal{N}$, instruction $I$, observation $o_t$
\State $\mathrm{mode}\gets\textsc{Navigator}$
\State $d\gets\textsc{Unknown}$
\While{episode is active}
  \If{$\mathrm{mode}=\textsc{Navigator}$}
    \State $d\gets\textsc{Unknown}$; $q\gets 0$
    \If{GroundingDINO detects a visible staircase}
      \State $(d,q)\gets\mathrm{VLM}(I,o_t)$
    \EndIf
    \If{$\bigl(d\in\{\textsc{Up},\textsc{Down}\}\ \lor\
        q\geq\delta_{\mathrm{act}}\bigr)$ and budget remains}
      \State initialize PACE context
      \State $\mathrm{mode}\gets\textsc{PACE}$
    \Else
      \State execute $\mathcal{N}(I,o_t)$ and acquire $o_{t+1}$
    \EndIf
  \Else
    \State $(\hat{\phi}_t,\hat{\tau}_t)\gets\mathrm{PACE}(c_t,d)$
    \If{$\hat{\phi}_t=\textsc{Exit}$}
      \State $\mathrm{mode}\gets\textsc{Navigator}$
    \Else
      \State execute $\hat{a}_{t,1}$ and acquire $o_{t+1}$
      \State update context $c_{t+1}$
      \If{episode completion or safety/budget event}
        \State $\mathrm{mode}\gets\textsc{Navigator}$
      \EndIf
    \EndIf
  \EndIf
\EndWhile
\end{algorithmic}
\end{algorithm}

\paragraph{Handoff and reactivation.}
PACE executes only the first primitive $\hat{a}_{t,1}$ of each
short-horizon proposal and replans from the subsequent RGB-D
observation. The terminal \texttt{STOP} ends only the current proposal
and triggers the next closed-loop decision; it neither terminates the
navigation episode nor immediately returns control to the navigator.
PACE terminates upon episode completion and otherwise hands control back
when the phase head predicts \textsc{Exit} or when a safety or budget
limit is reached. In simulation, each takeover is limited to 200
executed primitives and is interrupted after five consecutive forward
collisions or six consecutive forward attempts producing less than
$0.05\,\mathrm{m}$ displacement. After handoff, the navigator receives
the latest observation and may reactivate PACE through the same
staircase-detection and VLM-based activation gate, provided that
traversal remains incomplete and sufficient takeover budget remains.

Consequently, a handoff occurring before the stair exit has actually
been reached is caused by the phase head rather than by the
proposal-level \texttt{STOP}. Specifically, an \textsc{Exit} prediction
emitted while the agent is still traversing returns control immediately,
consistent with the phase-boundary ambiguity reported in the main
paper. The reactivation gate makes such premature handoffs recoverable:
if the staircase remains visible and the VLM again predicts a traversal
direction or confidently confirms its relevance, PACE can be
reactivated in both simulation and real-world deployment.

\section{Expanded Evaluation Results}

Table~\ref{tab:supp_full_ci} expands the compact significance markers
in the main paper by reporting the exact effect sizes and confidence
intervals for all 48 method--metric comparisons. PACE yields robust
gains in 9/12 SR, 12/12 OSR, 8/12 SPL, and 11/12 NDTW comparisons. Of
the remaining eight comparisons, seven are inconclusive; the only
robust decrease is NDTW for 3-step-Nav on RxR-CE-CF
($\Delta=-0.051$, 95\% CI $[-0.081,-0.022]$).

Table~\ref{tab:supp_takeover_ci} complements the takeover analysis in
the main paper with its exact SR intervals. All 12 intervals are
strictly above zero, with gains ranging from 9.68 to 44.12 percentage
points, providing consistent evidence across navigator--benchmark
pairs.

Alone and +PACE are aligned by episode ID but use independently sampled
VLM outputs. Thus, the analysis retains the realized end-to-end
variation rather than relying on matched VLM decisions. The uniformly
positive takeover-conditioned intervals indicate that the observed SR
advantage remains stable under this variation, although repeated
decoding seeds would be required to isolate VLM-induced variance.

\begin{table}[!htbp]
\centering

{\small
\setlength{\tabcolsep}{2pt}
\renewcommand{\arraystretch}{1.05}

\begin{adjustbox}{max width=\linewidth}
\begin{tabular}{@{}l
  c@{\hspace{0.4em}}c@{}r@{,\;}r@{}c
  c@{\hspace{0.4em}}c@{}r@{,\;}r@{}c
  @{}
}
\toprule

\multirow[c]{2}{*}{
  \raisebox{-0.6ex}[0pt][0pt]{Method}
}
& \multicolumn{5}{c}{R2R-CE-CF}
& \multicolumn{5}{c}{RxR-CE-CF} \\

\cmidrule(lr){2-6}
\cmidrule(lr){7-11}

& $\Delta$SR (pp)
& \multicolumn{4}{c}{95\% CI}
& $\Delta$SR (pp)
& \multicolumn{4}{c}{95\% CI} \\
\midrule

VLN-Zero
& 44.12 & [ & 34.31 & 53.92 & ]
& 33.33 & [ & 16.67 & 54.17 & ] \\

3-step-Nav
& 26.67 & [ & 16.30 & 36.30 & ]
& 9.68  & [ & 2.15  & 18.28 & ] \\

CA-Nav
& 16.96 & [ & 7.60  & 25.73 & ]
& 22.08 & [ & 10.39 & 33.77 & ] \\

STRIDER
& 27.78 & [ & 13.89 & 40.28 & ]
& 18.92 & [ & 8.11  & 32.43 & ] \\

Open-Nav
& 14.00 & [ & 3.00  & 25.00 & ]
& 16.39 & [ & 4.92  & 27.87 & ] \\

HSGM
& 29.00 & [ & 21.00 & 37.00 & ]
& 40.54 & [ & 30.63 & 50.45 & ] \\

\bottomrule
\end{tabular}
\end{adjustbox}
}

\caption{
Takeover-conditioned SR differences and their percentile
paired-bootstrap 95\% confidence intervals.
}
\label{tab:supp_takeover_ci}
\end{table}

\begin{table*}[!t]
\centering

{\small
\setlength{\tabcolsep}{1pt}
\renewcommand{\arraystretch}{1.05}

\resizebox{\linewidth}{!}{
\begin{tabular}{@{}l
  *{8}{c@{}r@{,}r@{}c}
  @{}
}
\toprule

\multirow[c]{2}{*}{
  \raisebox{-0.6ex}[0pt][0pt]{Method}
}
& \multicolumn{16}{c}{R2R-CE-CF}
& \multicolumn{16}{c}{RxR-CE-CF} \\

\cmidrule(lr){2-17}
\cmidrule(lr){18-33}

& \multicolumn{4}{c}{$\Delta$SR (pp)}
& \multicolumn{4}{c}{$\Delta$OSR (pp)}
& \multicolumn{4}{c}{$\Delta$SPL (pp)}
& \multicolumn{4}{c}{$\Delta$NDTW}
& \multicolumn{4}{c}{$\Delta$SR (pp)}
& \multicolumn{4}{c}{$\Delta$OSR (pp)}
& \multicolumn{4}{c}{$\Delta$SPL (pp)}
& \multicolumn{4}{c}{$\Delta$NDTW} \\
\midrule

\multirow[c]{2}{*}{VLN-Zero}
& \multicolumn{4}{c}{16.86}
& \multicolumn{4}{c}{19.92}
& \multicolumn{4}{c}{12.46}
& \multicolumn{4}{c}{0.219}
& \multicolumn{4}{c}{4.15}
& \multicolumn{4}{c}{4.15}
& \multicolumn{4}{c}{2.73}
& \multicolumn{4}{c}{0.035} \\
& [ & 11.88 & 22.22 & ]
& [ & 14.56 & 25.67 & ]
& [ & 8.13 & 16.88 & ]
& [ & 0.184 & 0.255 & ]
& [ & 0.92 & 7.37 & ]
& [ & 0.92 & 7.37 & ]
& [ & 0.66 & 5.10 & ]
& [ & 0.010 & 0.062 & ] \\
\addlinespace[1pt]

\multirow[c]{2}{*}{3-step-Nav}
& \multicolumn{4}{c}{12.64}
& \multicolumn{4}{c}{24.52}
& \multicolumn{4}{c}{6.58}
& \multicolumn{4}{c}{0.074}
& \multicolumn{4}{c}{4.15}
& \multicolumn{4}{c}{14.29}
& \multicolumn{4}{c}{-1.16}
& \multicolumn{4}{c}{-0.051} \\
& [ & 7.28 & 18.39 & ]
& [ & 18.01 & 31.03 & ]
& [ & 2.40 & 10.76 & ]
& [ & 0.041 & 0.108 & ]
& [ & 0.46 & 7.83 & ]
& [ & 9.22 & 19.35 & ]
& [ & -3.87 & 1.38 & ]
& [ & -0.081 & -0.022 & ] \\
\addlinespace[1pt]

\multirow[c]{2}{*}{CA-Nav}
& \multicolumn{4}{c}{10.73}
& \multicolumn{4}{c}{28.74}
& \multicolumn{4}{c}{4.69}
& \multicolumn{4}{c}{0.060}
& \multicolumn{4}{c}{7.83}
& \multicolumn{4}{c}{13.36}
& \multicolumn{4}{c}{3.41}
& \multicolumn{4}{c}{0.033} \\
& [ & 4.21 & 17.24 & ]
& [ & 21.84 & 36.02 & ]
& [ & 1.20 & 8.16 & ]
& [ & 0.030 & 0.091 & ]
& [ & 3.69 & 11.98 & ]
& [ & 8.76 & 17.97 & ]
& [ & 1.64 & 5.36 & ]
& [ & 0.014 & 0.053 & ] \\
\addlinespace[1pt]

\multirow[c]{2}{*}{STRIDER}
& \multicolumn{4}{c}{4.60}
& \multicolumn{4}{c}{7.66}
& \multicolumn{4}{c}{1.69}
& \multicolumn{4}{c}{0.034}
& \multicolumn{4}{c}{5.99}
& \multicolumn{4}{c}{7.83}
& \multicolumn{4}{c}{4.84}
& \multicolumn{4}{c}{0.031} \\
& [ & -0.38 & 9.96 & ]
& [ & 2.30 & 13.03 & ]
& [ & -2.64 & 6.06 & ]
& [ & 0.009 & 0.059 & ]
& [ & 3.23 & 9.22 & ]
& [ & 4.61 & 11.52 & ]
& [ & 2.39 & 7.54 & ]
& [ & 0.010 & 0.053 & ] \\
\addlinespace[1pt]

\multirow[c]{2}{*}{Open-Nav}
& \multicolumn{4}{c}{4.60}
& \multicolumn{4}{c}{13.03}
& \multicolumn{4}{c}{1.56}
& \multicolumn{4}{c}{0.053}
& \multicolumn{4}{c}{3.69}
& \multicolumn{4}{c}{5.99}
& \multicolumn{4}{c}{2.92}
& \multicolumn{4}{c}{0.065} \\
& [ & -0.77 & 9.96 & ]
& [ & 6.90 & 19.16 & ]
& [ & -3.22 & 6.25 & ]
& [ & 0.028 & 0.079 & ]
& [ & 0.00 & 7.37 & ]
& [ & 3.23 & 9.22 & ]
& [ & -0.34 & 6.31 & ]
& [ & 0.042 & 0.089 & ] \\
\addlinespace[1pt]

\multirow[c]{2}{*}{HSGM}
& \multicolumn{4}{c}{18.39}
& \multicolumn{4}{c}{20.31}
& \multicolumn{4}{c}{15.06}
& \multicolumn{4}{c}{0.053}
& \multicolumn{4}{c}{17.97}
& \multicolumn{4}{c}{23.96}
& \multicolumn{4}{c}{12.53}
& \multicolumn{4}{c}{0.081} \\
& [ & 11.49 & 25.29 & ]
& [ & 13.41 & 26.82 & ]
& [ & 9.75 & 20.25 & ]
& [ & 0.021 & 0.084 & ]
& [ & 11.52 & 24.42 & ]
& [ & 17.51 & 30.41 & ]
& [ & 7.93 & 17.19 & ]
& [ & 0.048 & 0.114 & ] \\

\bottomrule
\end{tabular}
}
}

\caption{
Observed differences $\Delta=+\mathrm{PACE}-\mathrm{Alone}$ under the
main-paper protocol. The first and second lines for each method report
$\Delta$ and its percentile paired-bootstrap 95\% CI, respectively,
using 10,000 episode-aligned resamples.
}
\label{tab:supp_full_ci}
\end{table*}

\section{Real-World Robot Deployment Details}

Table~\ref{tab:supp_robot_hardware} summarizes the real-world hardware
and computation assignment. The Unitree Go2 Edu carries a RealSense
D435i and a Livox Mid-360, while its Jetson Orin NX runs sensing,
motion control, and data-recording services. Learned-model inference is
performed offboard: PACE runs on an RTX 3070 Ti laptop, which reaches the
same self-hosted Qwen-3.6-27B-FP8 deployment used in simulation. As in
simulation, the high-level VLM is served locally on our own workstation
rather than by a third-party service; the laptop only acts as a client
and reaches that self-hosted endpoint over HTTPS within a private
network.

\begin{table}[!htbp]
\centering

{\small
\setlength{\tabcolsep}{2pt}
\renewcommand{\arraystretch}{1.03}

\begin{adjustbox}{max width=\linewidth}
\begin{tabular}{ll
  @{}
}
\toprule
Component & Configuration / role \\
\midrule
Robot & Unitree Go2 Edu \\
Onboard compute & Jetson Orin NX (16\,GB); sensing and motion \\
RGB-D sensor & RealSense D435i; $848{\times}480$ at 30\,Hz \\
LiDAR & Livox Mid-360; offline mapping only \\
Offboard runtime & RTX 3070 Ti laptop; PACE and VLM client \\
High-level VLM & self-hosted Qwen-3.6-27B-FP8; HTTPS client \\
Robot link & Authenticated HTTP over a private network \\
\bottomrule
\end{tabular}
\end{adjustbox}
}

\caption{
Hardware and computation assignment for real-world deployment.
}
\label{tab:supp_robot_hardware}
\end{table}

The offboard runtime requests fresh RGB-D observations and submits one
discrete action at a time to the NX through authenticated HTTP.
\texttt{FORWARD} is mapped to a $0.25\,\mathrm{m}$ relative target and
tracked from sport-mode odometry using velocity-reference MPC at
10\,Hz with a 15-step prediction horizon. \texttt{LEFT} and
\texttt{RIGHT} use odometry-closed-loop $30^\circ$ rotations. Velocity
commands are sent at 50\,Hz, and each primitive is allocated a
2\,s execution budget. Unitree's built-in obstacle avoidance is enabled
on level ground and disabled during stair traversal to avoid altering
the commanded stair motion. Together with 0.2--0.5\,s VLM or
approximately 0.2\,s PACE inference, the control cycle is 2.2--2.5\,s
under the reported deployment protocol.

\begin{figure}[!htbp]
\centering
\begin{minipage}{\linewidth}
\includegraphics[width=0.485\columnwidth]{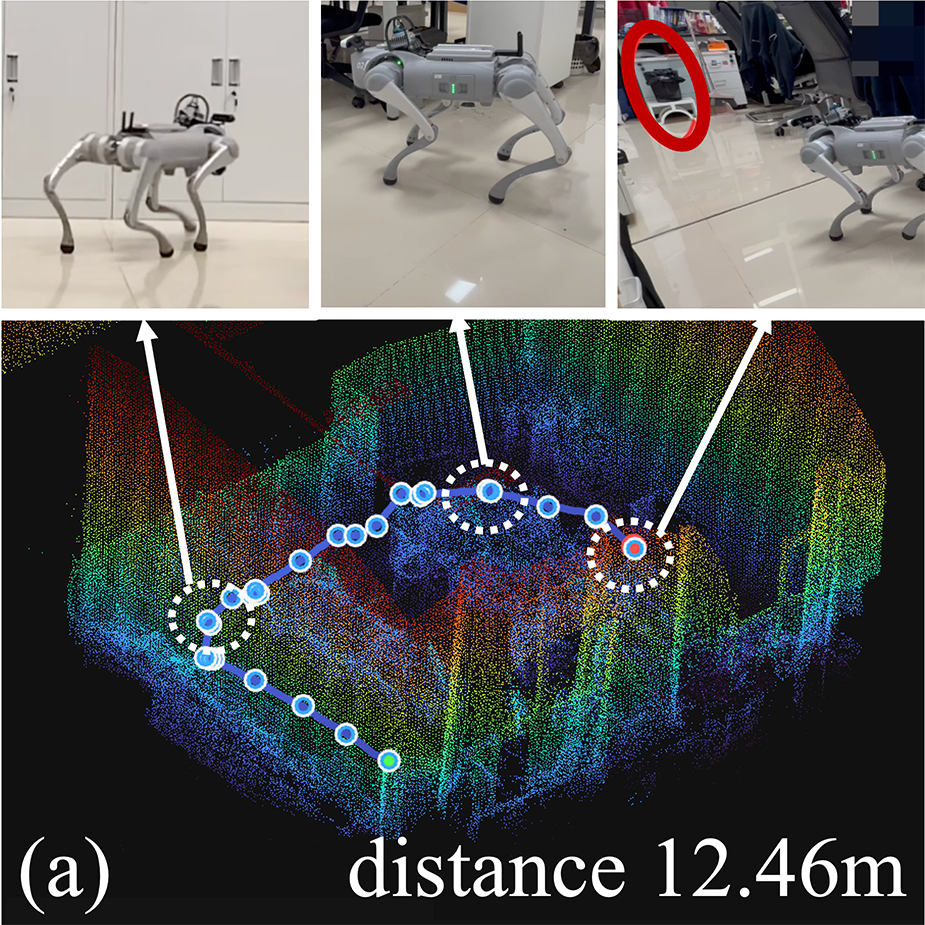}
\hfill
\includegraphics[width=0.485\columnwidth]{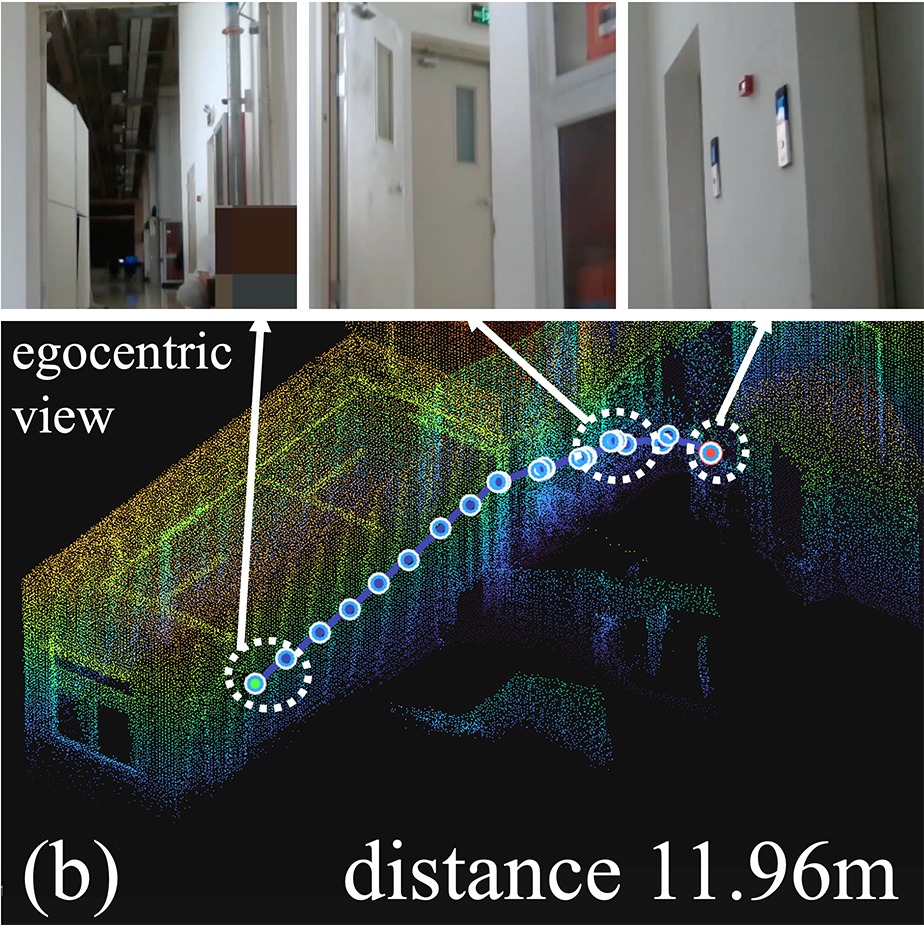}
\end{minipage}
\caption{
Additional real-world navigation examples.
(a) \textit{``Walk forward to the end, turn right, and continue
straight until you reach the trash bin.''}
(b) \textit{``Walk forward through the first doorway, turn right at
the second doorway, and stop in front of the elevator.''}
The point-cloud maps visualize the executed trajectories and are not
used for planning or as input to PACE.
}
\label{fig:supp_robot_examples}
\end{figure}

The Mid-360 measurements are processed offline for SLAM and trajectory
visualization and are not used for planning or as input to PACE. In the
unseen two-floor stairwell, the main observed failure modes were
left--right oscillation, reduced railing clearance, and a premature
handoff before the stair exit was actually reached. In the last case, the
navigator received the fresh observation, the VLM reassessed the
instruction as incomplete, and PACE was reactivated through the same
detection and direction gate, completing the traversal. These
observations expose remaining sim-to-real sensitivity to viewpoint shifts
and contact perturbations.
Figure~\ref{fig:supp_robot_examples} complements the stairwell case with
two additional unseen indoor navigation examples; their point-cloud
maps are visualization outputs only.

\section{Failure Case Analysis}

\noindent\textbf{Protocol.}
We pool the 261 R2R-CE-CF and 217 RxR-CE-CF val-unseen episodes, yielding
478 episode-aligned evaluations per baseline under the main-paper
protocol. Failed episodes are assigned to four mutually exclusive
categories in priority order: \emph{stair transition} when the aligned
PACE run invokes stair takeover on that episode, \emph{goal stopping}
when $\mathrm{OSR}=1$ but $\mathrm{SR}=0$, \emph{route deviation} when
$\mathrm{NDTW}<0.30$, and \emph{local execution} for the remaining
failures. Each segment is normalized by all 478 episodes rather than only
the failed ones, so the segments of one bar sum to that baseline's
overall failure rate.

Two properties of this decomposition should be noted. First, the
stair-transition category is defined by the takeover decision of the
aligned PACE run, so it identifies episodes whose failure coincides with
a stair transition rather than establishing that the stairs caused the
failure. Second, the priority order assigns an episode that both fails at
a stair transition and stops away from the goal to the stair-transition
category; the categories are therefore mutually exclusive by
construction but not independent.

\begin{figure}[!htbp]
\centering
\includegraphics[width=\linewidth]{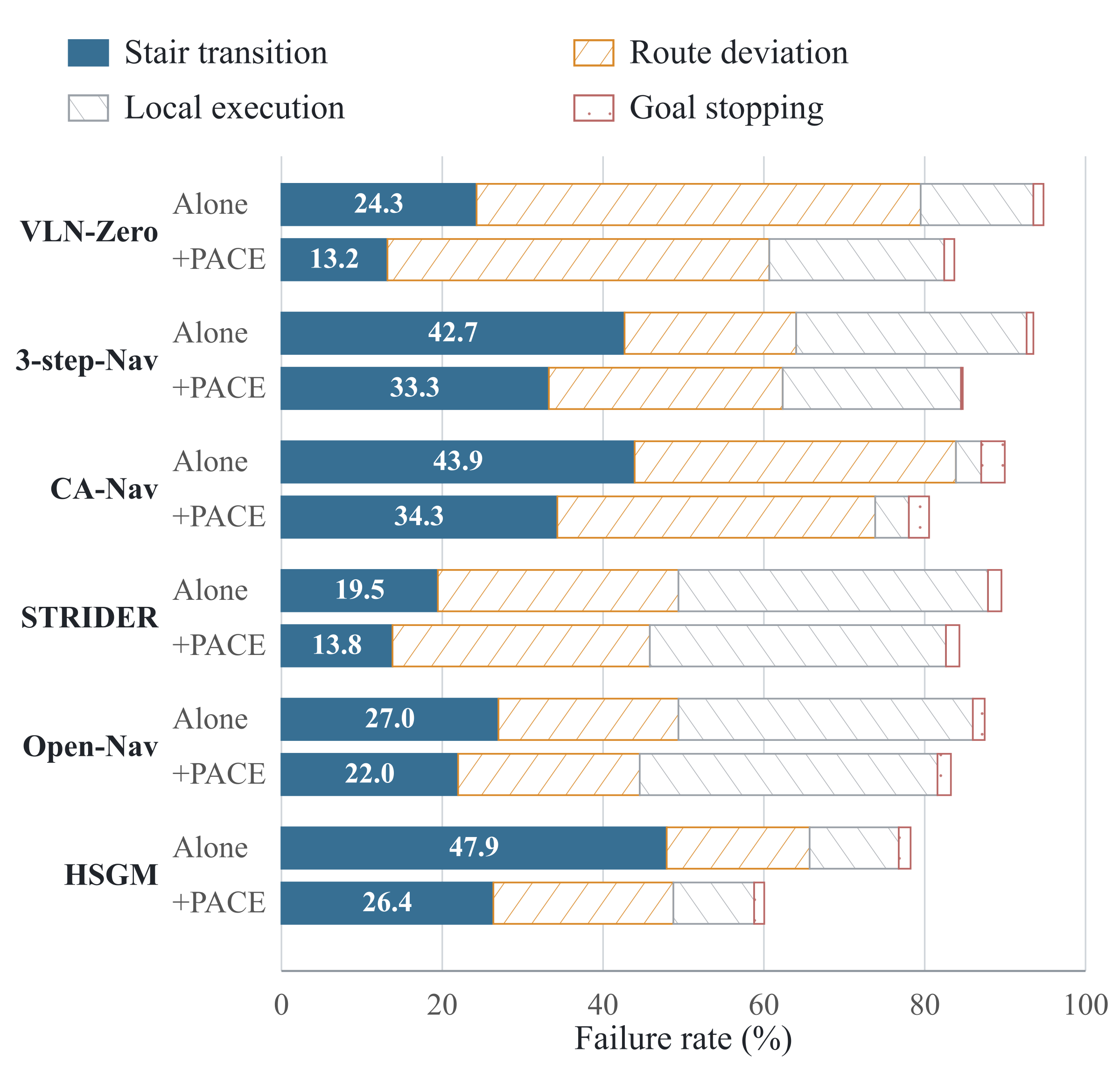}
\caption{
Failure-rate decomposition across six navigation baselines on the 478
episode-aligned val-unseen episodes. Segment lengths are normalized by
all episodes, such that each bar represents the overall failure rate.
Blue highlights stair-transition failures.
}
\label{fig:failure_reason_decomposition}
\end{figure}

\noindent\textbf{Results.}
As shown in Fig.~\ref{fig:failure_reason_decomposition}, PACE reduces
the average overall failure rate from $88.9\%$ to $79.4\%$
($-9.5$ percentage points) across all six baselines. The average
stair-transition failure rate decreases more substantially, from
$34.2\%$ to $23.8\%$ ($-10.4$ points, or $30.4\%$ relatively).
HSGM exhibits the largest improvement, with stair-related failures
decreasing from $47.9\%$ to $26.4\%$. Since the stair-transition
reduction of $10.4$ points exceeds the overall reduction of $9.5$ points,
the aggregate gain is attributable to the stair-transition category, with
the remaining categories contributing no net reduction. This is
consistent with PACE acting locally at constrained transitions rather
than improving global instruction following, and with the smaller NDTW
gains reported in the main paper.

\begin{figure*}[!p]
\centering
\includegraphics[width=0.82\textwidth]{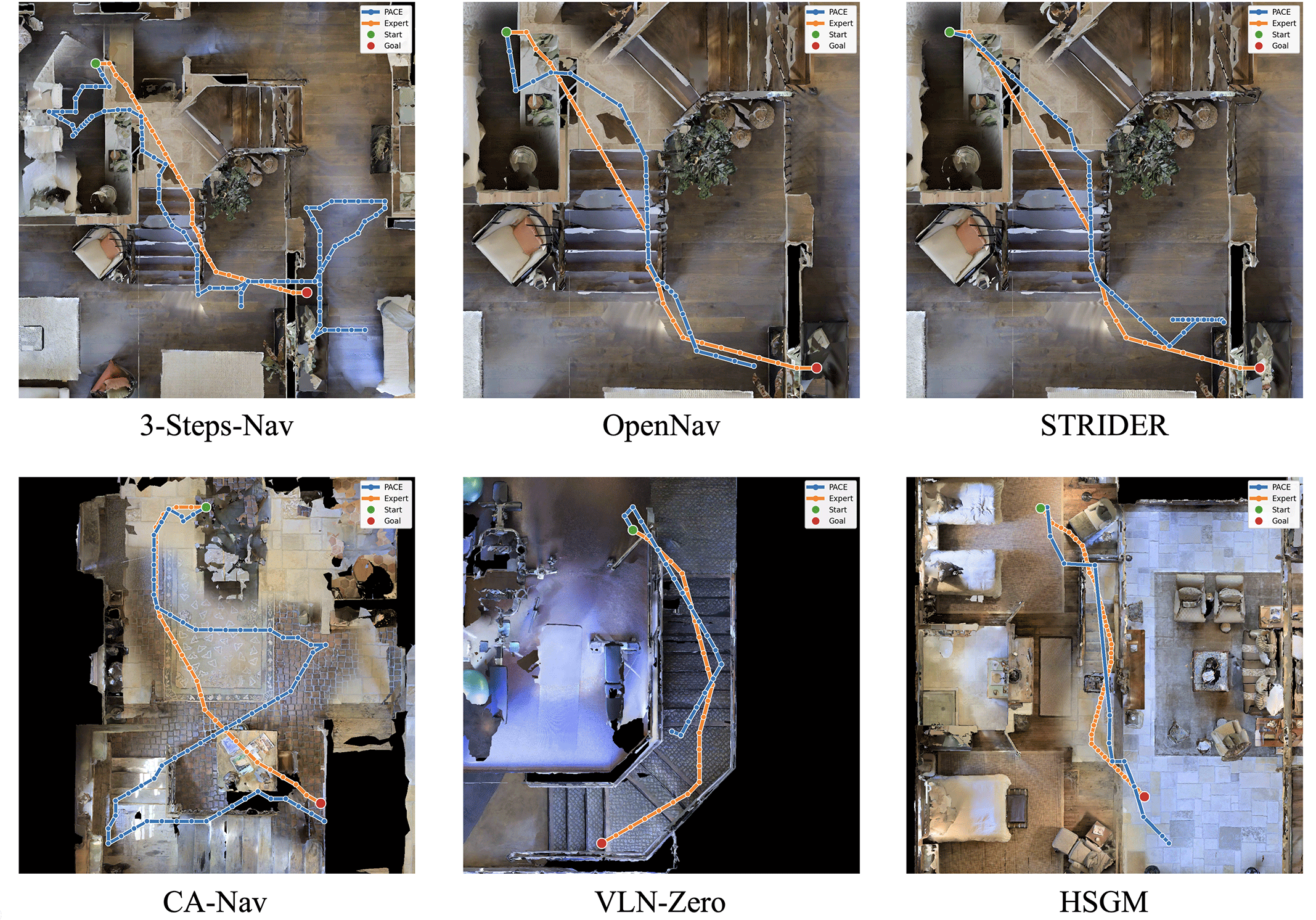}
\caption{
Representative successful navigation cases with PACE. The top row
compares 3-step-Nav, Open-Nav, and STRIDER in the same scene, showing
that PACE supports navigators with different internal mechanisms. The
bottom row presents additional cases across CA-Nav, VLN-Zero, and HSGM.
In particular, PACE completes the traversal when activated either on
the staircase (VLN-Zero) or near the stair entrance (HSGM). Blue and
orange denote the PACE-executed and expert trajectories, respectively.
}
\label{fig:supp_success_cases}
\end{figure*}

\paragraph{Qualitative success cases.}
Figure~\ref{fig:supp_success_cases} illustrates representative
successful trajectories across all six navigators. In the same scene,
PACE enables 3-step-Nav, Open-Nav, and STRIDER to complete the
cross-floor route despite their different perception, memory, and
planning mechanisms. The remaining examples further demonstrate
robustness to takeover timing: PACE can guide the agent through the
transition when activated either after it has entered the staircase, as
in VLN-Zero, or near the stair entrance, as in HSGM.

\begin{figure*}[!p]
\centering
\includegraphics[width=0.82\textwidth]{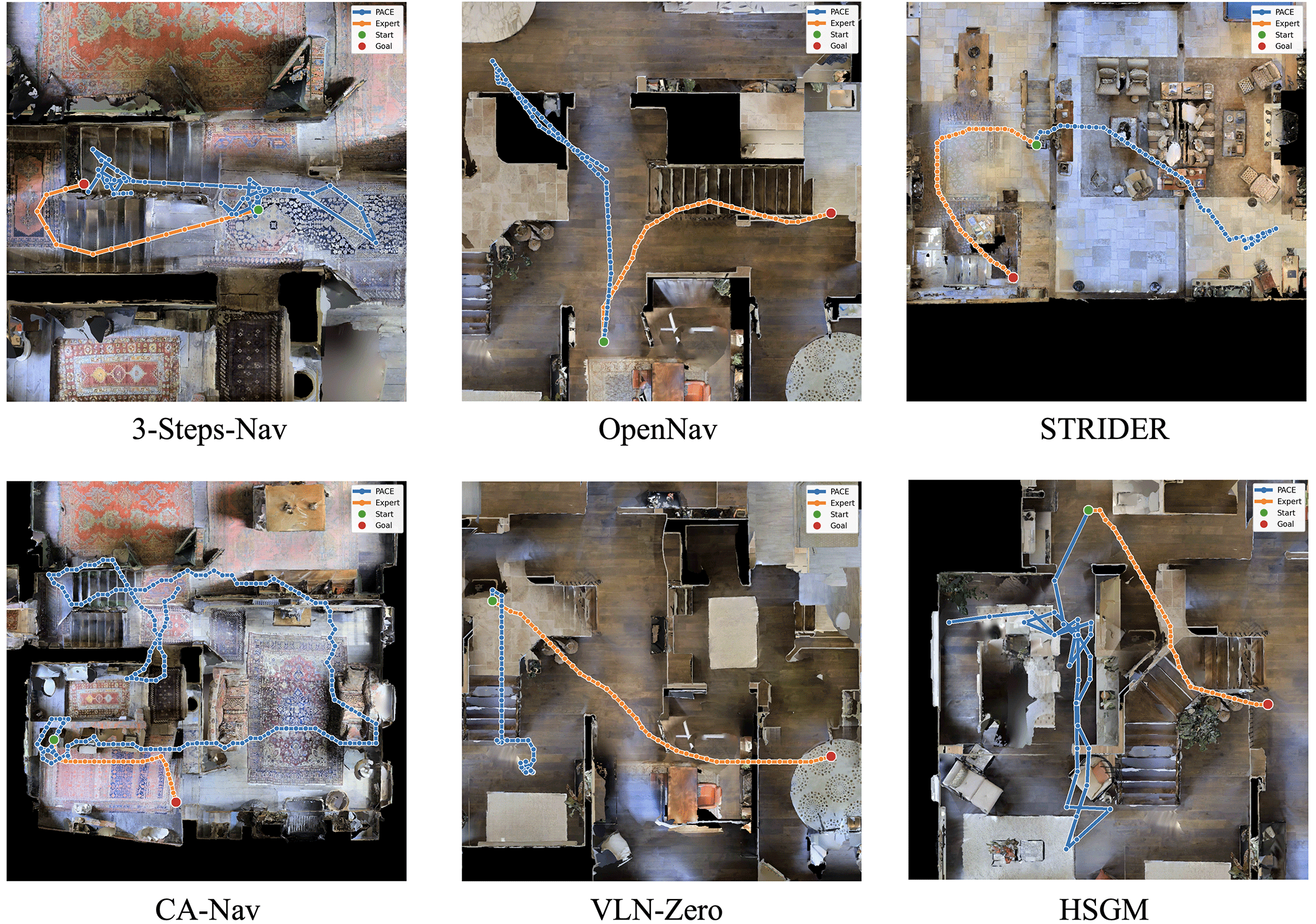}
\caption{
Representative failure cases across the six navigators. Blue and
orange denote the executed and expert trajectories, respectively.
Failures arise from both PACE-specific execution errors and limitations
of the underlying navigators.
}
\label{fig:supp_failure_cases}
\end{figure*}

\paragraph{Failure cases.}
Figure~\ref{fig:supp_failure_cases} illustrates several remaining
failure modes. For CA-Nav, PACE fails to exit and return control,
exhausts its takeover budget, and drives the agent away from the target,
constituting a harmful takeover. With 3-step-Nav, repeated collisions
lead to persistent turning without effective recovery. For Open-Nav and
STRIDER, the underlying navigators fail to identify a reasonable global
direction, which cannot be corrected by a local execution module. In
the HSGM case, PACE is never activated, so the failure originates from
the baseline navigation process rather than PACE execution. These cases
highlight limitations in exit prediction, collision recovery, upstream
direction selection, and takeover triggering.

\end{document}